 \documentclass[accepted]{uai2022} 

\usepackage[british]{babel}

\usepackage{natbib} 
\usepackage[dvipsnames]{xcolor}
\usepackage{pifont}
\usepackage{booktabs} 
\usepackage{tikz} 
\usepackage{amsmath}
\usepackage{amssymb}
\usepackage{mathtools}
\usepackage{amsthm}
\usepackage{placeins}
\usepackage{multirow}
\usepackage{adjustbox}
\usepackage{tabularx}
\usepackage{color, colortbl}
\definecolor{Gray}{gray}{0.9}
\usepackage{subcaption}

\newcommand{\x}{\boldsymbol{x}}
\newcommand{\y}{\boldsymbol{y}}
\newcommand{\xo}{\x^{\mathrm{o}}}
\newcommand{\xu}{\x^{\mathrm{u}}}
\newcommand{\yo}{\y^{\mathrm{o}}}
\newcommand{\yu}{\y^{\mathrm{u}}}
\newcommand{\Xo}{X^{\mathrm{o}}}
\newcommand{\Xu}{X^{\mathrm{u}}}
\newcommand{\Yo}{Y^{\mathrm{o}}}
\newcommand{\Yu}{Y^{\mathrm{u}}}
\newcommand{\z}{\boldsymbol{z}}
\newcommand{\m}{\boldsymbol{m}}
\newcommand{\s}{\boldsymbol{s}}

\newcommand{\f}{\boldsymbol{f}}

\renewcommand{\L}{\mathcal{L}}
\newcommand{\Y}{\mathcal{Y}}
\newcommand{\N}{\mathcal{N}}
\newcommand{\GP}{\mathcal{GP}}

\newcommand{\X}{\mathcal{X}}

\newcommand{\Z}{\mathcal{Z}}

\newcommand{\E}{\mathbb{E}}
\newcommand{\R}{\mathbb{R}}

\renewcommand{\u}{\boldsymbol{u}}
\newcommand{\tr}{\text{tr}}
\newcommand{\cmark}{\color{ForestGreen}\ding{51}}%
\newcommand{\xmark}{\color{RubineRed}\ding{55}}
\DeclarePairedDelimiterX{\infdivx}[2]{(}{)}{%
  #1\;\delimsize\|\;#2%
}

\title{Learning Conditional Variational Autoencoders with Missing Covariates}

\author[1]{\href{mailto:<siddharth.ramchandran@aalto.fi>}{Siddharth Ramchandran}{}}
\author[2]{Gleb Tikhonov}
\author[1]{Otto L\"onnroth}
\author[3]{Pekka Tiikkainen}
\author[1]{Harri L\"ahdesm\"aki}
\affil[1]{%
    Department of Computer Science\\
    Aalto University\\
    Espoo, Finland
}
\affil[2]{%
    Organismal and Evolutionary Biology Research Programme\\
    University of Helsinki\\
    Helsinki, Finland
}
\affil[3]{%
    Bayer Pharmaceuticals\\
    Espoo, Finland
  }
  
\begin{document}
\maketitle

\begin{abstract}
Conditional variational autoencoders (CVAEs) are versatile deep generative models that extend the standard VAE framework by conditioning the generative model with auxiliary covariates. The original CVAE model assumes that the data samples are independent, whereas more recent conditional VAE models, such as the Gaussian process (GP) prior VAEs, can account for complex correlation structures across all data samples. While several methods have been proposed to learn standard VAEs from partially observed datasets, these methods fall short for conditional VAEs. In this work, we propose a method to learn conditional VAEs from datasets in which auxiliary covariates can contain missing values as well. The proposed method augments the conditional VAEs with a prior distribution for the missing covariates and estimates their posterior using amortised variational inference. At training time, our method marginalises the uncertainty associated with the missing covariates while simultaneously maximising the evidence lower bound. We develop computationally efficient methods to learn CVAEs and GP prior VAEs that are compatible with mini-batching. Our experiments on simulated datasets as well as on a clinical trial study show that the proposed method outperforms previous methods in learning conditional VAEs from non-temporal, temporal, and longitudinal datasets.
\end{abstract}

\section{Introduction}
\label{sec:intro}
\citet{kingma2013auto} introduced the auto-encoding variational Bayes (AEVB) as a general technique for efficient variational inference. 
A very popular instance of this approach is the variational autoencoder (VAE) \citep{kingma2013auto, rezende2014stochastic}. VAEs have been extensively used in representation learning in order to learn low-dimensional manifolds of highly complex data \citep{higgins2017beta, kulkarni2015deep, fortuin2018som}, impute missing values in data \citep{ma2019eddi, ramchandran2021longitudinal, fortuin2019multivariate, nazabal2018handling}, and to learn the underlying generative model \citep{zhao2019infovae, germain2015made, mita2021identifiable, vedantam2018generative}. Given the flexibility of VAEs, they have recently become highly popular in a variety of fields, including e.g.,\ molecular biology, chemical design, natural language generation, astronomy (see e.g.,\ \citep{kingma2019introduction}) and many more. 

Conditional VAEs (CVAEs) \citep{sohn2015learning} were proposed as an extension to the VAE that models the distribution of the high-dimensional output as a generative model conditioned on auxiliary covariates (control variables). The CVAE model offers an approach to control the data generating process of a VAE and thereby perform structured output predictions. However, though the standard VAEs and CVAEs are powerful generative models for complex datasets, they ignore the possible correlations between the data samples. Recent work has extended the VAE modelling framework to incorporate arbitrary correlations across all data samples by replacing the i.i.d. standard Gaussian prior with a Gaussian process (GP) prior \citep{casale2018gaussian, fortuin2019multivariate, ramchandran2021longitudinal}. These GP prior VAEs are also conditional generative models as the generative process depends on auxiliary covariates (such as time, image rotation, etc.). GP prior VAEs have been demonstrated to be especially well suited for temporal and longitudinal datasets.

Many real-world datasets, such as clinical studies, usually contain a significant number of missing values. Moreover, real-world datasets contain variables that can be regarded as either covariates or dependent variables, both with varying amounts of missing information. For example, in a clinical study, dependent variables can comprise of lab measurements of a participant's blood sample, whereas covariates may contain information about the participant, such as age, sex, height, etc. Recent papers have demonstrated the ability of VAEs to handle missing values and have shown good imputation performance for the dependent variables. However, these models either do not incorporate the auxiliary variables or do not account for the missing values in the auxiliary variables. In other words, no methods have been developed to learn CVAEs with missing auxiliary variables.

\begin{figure}[t]
\centering
  \includegraphics[width=\linewidth]{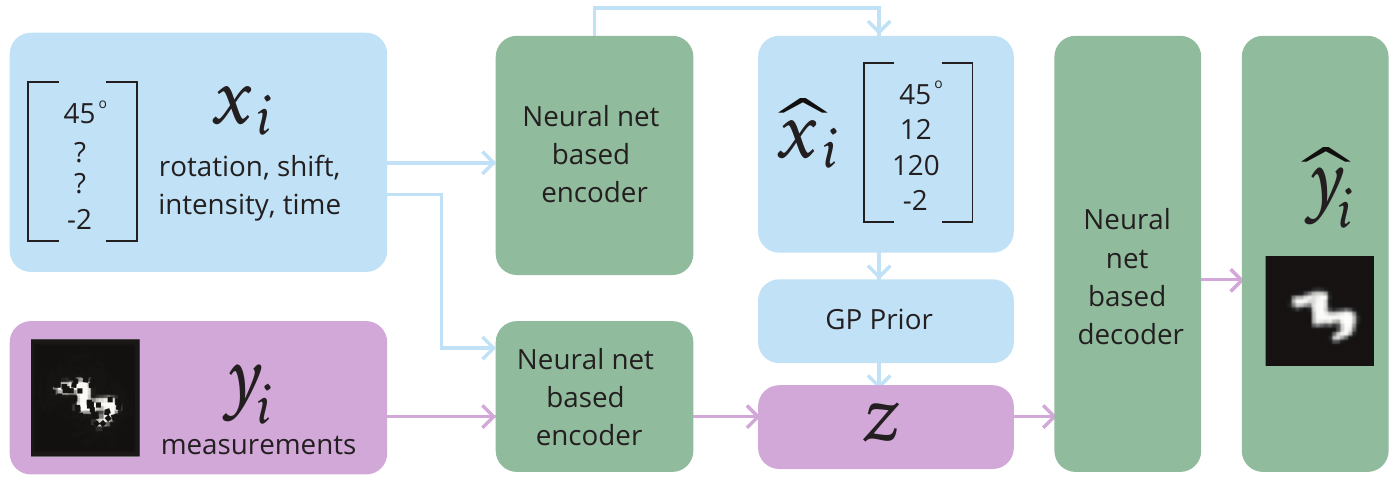}
  \caption{An overview of our method, applied to a GP prior VAE model. $\hat{x}_i$ and $\hat{y}_i$  refers to $x_i$ and $y_i$ respectively with all the missing values imputed.}
  \label{fig:overview}
\end{figure}

\paragraph{Contributions} In this work, we propose a probabilistic method to learn conditional VAE models from partially observed datasets that can contain missing values also in the auxiliary covariates. Fig. \ref{fig:overview} provides an overview of our method when applied to a GP prior VAE model. Our contributions can be summarised as follows:
\begin{itemize}
    \item We introduce an amortised variational distribution for learning the missing auxiliary covariates and propose a method that maximises the evidence lower bound objective while simultaneously marginalising uncertainty associated with the missing covariates.
    \item We develop a computationally efficient and mini-batching compatible method to learn GP prior VAEs from data that contain missing auxiliary covariates.
    \item We demonstrate the state-of-the-art performance of our method in learning CVAEs and GP prior VAEs from non-temporal, temporal, and longitudinal data.
\end{itemize}
The source code will be made available upon publication.

\section{Related Work}
\label{sec:related_work}
Various methods have been proposed to replace missing values with a substitute value that tries to approximate the true value. A standard approach is to replace the missing values of an attribute with the mean of the attribute \citep{richman2009missing}. Another popular approach is to approximate the missing values by observing the corresponding values of the instance's k-nearest neighbours (k-NN) \citep{richman2009missing}. However, such rule-based baseline approaches can be too simple for most complex datasets. 

Variational autoencoders (VAEs) \citep{kingma2013auto, rezende2014stochastic} are popular deep generative models that are used to model large and complex datasets by learning complex encoder and decoder neural networks that can map high-dimensional data to a low-dimensional space and vice-versa. However, most VAE models assume that the data is fully observed or choose the heuristic of substituting the unobserved values with zero for the encoder input \citep{nazabal2020handling, mattei2019miwae}. \citet{nazabal2020handling} obtained an ELBO which depends only on the observed values by integrating out the missing elements from the VAE objective. A similar approach was followed by \citet{mattei2019miwae} in which they adapted the importance weighted autoencoder proposed by \citet{burda2015importance} to missing data. \citet{collier2020vaes} developed a model of a corruption process which generates missing data and derived a tractable ELBO. Furthermore, \citet{ipsen2020not} proposed a variant of this model in which the missing process is dependent on the missing data. \citet{vedantam2018generative} provided a different perspective by proposing a VAE-based model that does not focus on the individual measurements (or pixels) that are observed, but on whether certain high-level attributes were observed. They make use of the high-level attributes to generate measurements (or pixels).

Conditional VAEs \citep{sohn2015learning} directly incorporate the auxiliary covariate information in the inference and generative networks. However, these models do not capture the subject-specific temporal structure and hence, are not well suited to temporal or longitudinal datasets. Gaussian process prior VAEs have been proposed as an extension to the conditional generative model that incorporate temporal correlations as well as external covariates via GP priors. The GPPVAE \citep{casale2018gaussian}, GP-VAE \citep{fortuin2019multivariate}, and L-VAE \citep{ramchandran2021longitudinal} are some of the GP prior VAE methods that have been proposed to impute the missing values in the data (or measurements). Deep generative modelling based semi-supervised learning approaches \citep{tulyakov2017hybrid, maaloe2016auxiliary} are an alternative modelling approach that treats the auxiliary covariates as instance labels. However, these methods do not focus on missing value imputation.

Though the majority of the above-mentioned methods allow for imputing the missing values in the data, none of them are designed to robustly handle the missing values that may be present in the auxiliary covariates. In this work, we propose a novel learning approach that simultaneously marginalises the missing values in the auxiliary covariates while achieving improved imputation performance in the partially observed data. Supplementary Tables~\ref{table:comparison} and~\ref{table:comparison_learning} contrast the features of our proposed method to the key related conditional VAE models and learning methods, respectively.

\section{Methods}
\label{sec:methods}
\paragraph{Problem setting}
We let $(\x, \y)$ to denote a single data sample, where $\y \in \Y = \R^D$ represent the $D$-dimensional observed (dependent) variables and $\x \in \X = \X_1 \times \ldots \times \X_Q$ represent the $Q$-dimensional auxiliary (independent) covariates. Covariates can be both discrete and continuous-valued with the domain of the $q^\text{th}$ covariate specified by $\X_q$. 

A {\em non-temporal dataset} of size $N$ is denoted as $X = (\x_1,\ldots, \x_N)^T$ and $Y = (\y_{1}, \ldots, \y_N)^T$. For example, in a cross-sectional biomedical application each pair $(\x_i,\y_i)$ could represent an individual patient

A {\em temporal dataset} is denoted similarly with the exception that a dataset $(X,Y)$ represents a $N$-length time-series for an individual patient.

A {\em longitudinal dataset} consists of $P$ instances (i.e., $P$ individual patients) 
and each instance $p$ is observed over $n_p$ irregularly sampled time points, $X_p = (\x_1^p, \ldots, \x_{n_p}^p)^T$ and $Y_p = (\y_1^p, \ldots, \y_{n_p}^p)^T$. This results in a total of $N = \sum_{p=1}^P n_p$ data samples, $X = (X_1^T, \ldots, X_P^T)^T = (\x_1,\ldots,\x_N)^T$, and $Y = (Y_1^T, \ldots, Y_P^T)^T =  (\y_1,\ldots,\y_N)^T$. 

A {\em latent embedding} for a single data sample is denoted as $\z_i \in \Z = \R^L$, whereas a latent embedding for a dataset is $Z = (\z_1,\ldots,\z_N)^T \in \mathbb{R}^{N \times L}$. For GP prior VAE models, we also index $Z$ across $L$ dimensions as $Z = (\bar{\z}_{1},\ldots,\bar{\z}_{L})$, where $\bar{\z}_{l} = (z_{1l},\ldots,z_{Nl})^T$ is a vector that contains the $l^{\mathrm{th}}$ dimension of the latent embedding for all $N$ samples.

We assume that for each data sample $(\x,\y)$ any subset of the covariates and observed variables may be missing completely at random. For the $i^\text{th}$ data sample, the observed parts of the data are denoted as $\xo_i$ and $\yo_i$, and the unobserved data as $\xu_i$ and $\yu_i$.
To simplify notation, we do not explicitly specify (using index or mask vectors) which dimensions of $\x_i$ and $\y_i$ are missing for each $i$, but assume that that information is included in $\xo_i$, $\xu_i$, $\yo_i$, and $\yu_i$. The observed and unobserved datasets are defined similarly as $\Xo$, $\Yo$, $\Xu$, and $\Yu$, where e.g.\ $\Yo = \{\yo_1,\ldots,\yo_N\}$.

\subsection{Variational Autoencoders}
\label{subsec:vae}
A latent variable generative model typically has the form of $p_{\omega}(\y, \z) = p_{\psi}(\y|\z)p_{\theta}(\z)$ being parameterised by $\omega = \{\psi, \theta\}$. The objective is to infer the latent variable $\z$ given $\y$. The posterior distribution $p_{\omega}(\z|\y) = p_{\psi}(\y|\z)p_{\theta}(\z)/p_{\omega}(\y)$, however, is usually intractable due to the lack of a closed-form marginalisation over the latent space. The VAE is a type of auto-encoding variational Bayes (AEVB) model which comprises of the generative model (also called probabilistic decoder) as well as an additional inference model (also called probabilistic encoder) 
$q_{\phi}(\z|\y)$ that approximates the true posterior. Making use of amortised variational inference (VI) that exploits the inference model $q_{\phi}(\z|\y)$ to obtain approximate distributions for each $\z_i$, the approximate inference problem is fitted by maximising the evidence lower bound (ELBO) of the marginal log-likelihood w.r.t.\ $\phi$:
\begin{align} \label{eq:ELBOVAE}
    &\log p_\omega(Y) \geq \mathcal{L}(\phi,\psi,\theta;Y) \\ 
    &\triangleq \mathbb{E}_{q_{\phi}}[\log p_{\psi}(Y|Z)]- \mathrm{KL}[q_{\phi}(Z|Y)||p_\theta(Z)] \rightarrow \max_\phi, \nonumber
\end{align}
where $\mathrm{KL}$ denotes the Kullback-Leibler (KL) divergence. In applications, both the encoder and decoder are typically parameterised by deep neural networks (DNNs). Thus, the generative model's parameters are typically learnt alongside the approximate inference by solving the joint optimisation problem $\mathcal{L}(\phi,\psi,\theta;Y)\rightarrow \max_{\phi,\psi}$. The standard VAE assumes that the generative model factorises across the samples as $p_{\omega}(Y,Z)=p_{\psi}(Y|Z)p_{\theta}(Z) = \prod_{i=1}^{N}p_{\psi}(\y_i|\z_i)p_{\theta}(\z_i)$. Combined with an encoder that factorises across the samples $q_{\phi}(Z|Y)=\prod_{i=1}^{N}q_{\phi}(\z_i|Y)$, the ELBO in eq.~\ref{eq:ELBOVAE} also factorises across the samples. Therefore, it is straightforward to apply computationally efficient mini-batch based stochastic gradient descent (SGD). In this work, we assume that the decoder has a Gaussian likelihood with a free variance parameter for each of the $D$ dimensions.
\subsection{Conditional VAE}
\label{subsec:cvae}
CVAEs extend the standard VAE framework by conditioning the generative model with auxiliary covariates. In a general form, the joint conditional distribution is written as  $p_{\omega}(\y, \z | \x) = p_{\psi}(\y | \z, \x)p_{\theta}(\z | \x)$ but in applications the prior is often simplified to $p_{\theta}(\z | \x) = p_{\theta}(\z)$. Similarly for the inference model, CVAEs implement an encoder network that is augmented by the auxiliary covariates, $q_{\phi}(\z | \y,\x)$.
The variational evidence lower bound objective for a CVAE can be obtained by conditioning the probabilities in eq.~\ref{eq:ELBOVAE} with the covariates $\x$, $\log p_\omega(Y | X) \geq \mathcal{L}(\phi,\psi,\theta;Y,X)$ \citep{sohn2015learning}. The ELBO for the standard CVAE is also directly amenable to mini-batched based SGD optimisation because the prior $p_{\theta}(\z | \x)$ (or $p_{\theta}(\z)$) is independent across the data samples. 

\subsection{Gaussian process prior VAE}
\label{subsec:gpvae}
The standard CVAE can neither capture any correlation structure across the data samples nor exploit that structure while making predictions as it assumes that the joint distribution factorises across the samples. One approach to extend the conditional generative models to tackle this would be to combine the power of VAEs with the ability to model correlations afforded by GPs. Models such as the GPPVAE \citep{casale2018gaussian}, GP-VAE \citep{fortuin2019multivariate}, and L-VAE \citep{ramchandran2021longitudinal} have been proposed to model multivariate time-series data by assuming a GP prior on the latent space $Z$. 

The key distinction between CVAE and GP prior VAEs is that the factorisable conditional prior $p_{\theta}(Z|X) = \prod_{i=1}^N p_{\theta}(\z_i|\x_i)$ is replaced by a GP prior. Specifically, consider a function $\f : \X \rightarrow \Z$ that maps the auxiliary covariates to the $L$-dimensional latent space, and denote $\z = \f(\x) = (f_1(\x),\ldots,f_L(\x))^T$. The GP prior VAEs assume that each dimension $l$ of the latent embedding $\z$ has a GP prior 
\begin{equation}
    f_l(\x) \sim \GP(\mu_l(\x),k_l(\x,\x'|\theta_l)),
\end{equation}
where $\mu_l(\x)$ is the mean (which we assume as $0$), $k_l(\x,\x'| \theta_l)$ is the covariance function which defines the covariance between any pair of auxiliary covariates $\x$ and $\x'$, and $\theta_l$ denotes the parameters of the covariance function. For any finite set of inputs $X$, the GP prior implies a joint multivariate Gaussian distribution for the function values $\bar{\z}_l = f_l(X) = (f_l(\x_1),\ldots,f_l(\x_N))^T$
\begin{equation}\label{eq:GPnormal}
    p_{\theta}(\bar{\z}_l|X) =  p_{\theta}(f_l(X)) = \N(\bar{\z}_l | \mathbf{0},K_{XX}^{(l)}),
\end{equation}
where $K_{XX}^{(l)}$ is the $N \times N$ covariance matrix with entries $\{K_{XX}^{(l)}\}_{i,j} = k_l(\x_i,\x_j | \theta_l)$. 

Note that while the GP priors for the latent dimensions are assumed to be apriori independent, the probabilistic decoder $p_{\psi}(Y|Z)$ can introduce arbitrary correlations between the dimensions. Therefore, the joint conditional prior can be written as 
\begin{align}
    P_{\theta}(Z|X) &= \prod_{l=1}^L p_{\theta}(\bar{\z}_l|X) = \prod_{l=1}^L \N(\bar{\z}_l | \mathbf{0}, K_{XX}^{(l)}). \label{eq:GP-prior-factorizes}
\end{align}

The choice of covariance functions for the latent GPs  are the fundamental differentiating factor among the various GP prior VAE models. While the early works (e.g.,~GPPVAE and GP-VAE) assumed very specific and restricted covariance functions, more recent works have proposed flexible and expressive covariance functions. Here, we will follow the L-VAE model~\citep{ramchandran2021longitudinal} that assumes additive covariance functions
\begin{equation} \label{eq:additive-kernel}
    k_l(\x,\x' | \theta_l) = \sum_{r=1}^R k_{l,r}(\x,\x' | \theta_{l,r}) + \sigma_{zl}^2,
\end{equation}
which implies $K_{XX}^{(l)} = \sum_{r=1}^R K_{XX}^{(l,r)} + \sigma_{zl}^2 I_N$. Other GP prior VAE models can be seen as special cases of the L-VAE model. The specific choice of the kernels $k_{l,r}$ depends on the application. 
We consider three different variants of the L-VAE model as our general GP prior VAE models: 

\paragraph{Regression GP prior VAEs} Here we assume that $R=1$, covariates are split into continuous and discrete-valued variables $\x = (\x^{\text{c}},\x^{\text{d}})$, and $k_l(\x,\x^{\prime}|\theta_l) = k_{l}^{\text{se}}(\x^{\text{c}},\x^{\text{c}\prime }|\theta_l) k_l^{\text{ca}}(\x^{\text{d}},\x^{\text{d}\prime}|\theta_l)$, where $k_l^{\text{se}}(\x^{\text{c}},\x^{\text{c}\prime }|\theta_l)$ is the squared exponential covariance function that depends on the continuous covariates $\x^{\text{c}}$ and $k_l^{\text{ca}}(\x^{\text{d}},\x^{\text{d}\prime}|\theta_l)$ is the product of categorical covariance functions, one for each categorical covariate within $\x^{\text{d}}$. To the best of our knowledge, regression GP prior VAEs have not been proposed in literature but these are general models for  high-dimensional data. 

\paragraph{Temporal GP prior VAEs} These models are defined similarly as regression GP prior VAEs with an exception that covariates are assumed to contain the time covariate, which makes these models particularly suitable for time-series data. 

\paragraph{Longitudinal GP prior VAEs} These models are specifically designed for high-dimensional longitudinal study designs, where data contains repeated measurements of each unique instance (e.g.,\ individual patient). We use the same model specification as in L-VAE \citep{ramchandran2021longitudinal}, where each additive covariance function $k_{l,r}(\x,\x'|\theta_{l,r})$ in eq.~\ref{eq:additive-kernel} depends only on a subset (e.g.,\ a pair) of covariates, and $k_{l,R}(\x,\x'|\theta_{l,R})$ accounts for the instance-specific effects and is defined as the product of categorical covariance function (for covariate that specifies the instance) and squared exponential covariance function (for time or age covariate). 

In GP prior VAEs models, the ELBO can be written as for the vanilla CVAE model in Section~\ref{subsec:cvae} with the exception that the prior does not factorise across the samples 
\begin{align} 
    &\log p_\omega(Y | X) \geq \mathcal{L}(\phi,\psi,\theta;Y, X) \nonumber \\
    &\triangleq \mathbb{E}_{q_{\phi}}[\log p_{\psi}(Y | Z)] - \mathrm{KL}[q_{\phi}(Z | Y)||p_\theta(Z | X)], \label{eq:ELBO_GPVAE2}  
\end{align}
where $p_\theta(Z | X)$ is the $L$-dimensional GP prior over the latent space and optimisation is done w.r.t.\ $\phi$, $\psi$, and $\theta$. 
The key differentiating factor among the various GP prior VAE models concerns how an approximation to the GPs is implemented and, thus, how the optimisation of the ELBO in eq.~\ref{eq:ELBO_GPVAE2} scales to big data. For example, GPPVAE makes use of a pseudo mini-batch approach that exploits the Taylor series expansion of the KL divergence, GP-VAE fits a separate time-series for each individual, and L-VAE proposes a more general mini-batching based approach that exploits the kernel structure of the GP prior. We will develop the optimisation procedure after augmenting the model to missing covariates. 

\subsection{Learning with missing covariates}
\label{subsec:elbo_cvae}
Here we describe a novel method to learn conditional VAE models from datasets that can contain missing values in both data $\y$ as well as auxiliary covariates $\x$. 
We shall first describe a general learning method for conditional VAEs and then proceed to describe how this can be used in specific conditional VAE models, CVAEs and GP prior VAEs.

In our problem setting, any subset of the covariates may be missing for each data sample. To account for the uncertainty in the missing covariates, we augment the generative model with a prior distribution for $\x$, $p_{\lambda}(\x)$, parameterised by ${\lambda}$. This distribution represents a distribution from which the covariates originate from.
The joint distribution of the augmented conditional model has the following form: $p_{\omega}(\y,\z,\x) = p_{\psi}(\y | \z) p_{\theta}(\z | \x) p_{\lambda}(\x)$, where the parameters of the generative model are extended accordingly $\omega = \{\psi,\theta, \lambda\}$.

Contrary to the commonly used standard normal prior for latent variables in VAEs, $p_{\lambda}(\x)$ can be also informative. For example, in biomedical applications covariates may include e.g.,\ the height of individual patients, which may be defined based on prior information about a study population.  In an alternative data-driven approach that we will use in this study, the parameters $\lambda$ of the prior can also be learned from data. We assume that the prior factorises across the dimensions, and that the continuous-valued covariates have a Gaussian distribution and the discrete-valued covariates have a categorical distribution, although our method is conceptually not limited to these assumptions. 

Given the observed dataset $(\Xo,\Yo)$, our goal is to approximate the true posterior distribution of the unobserved variables $Z$ and $\Xu$, $p_{\omega}(Z,\Xu | \Yo, \Xo)$, using amortised variational inference. We use a conditionally independent, factorisable variational approximation
\begin{align*}
q_{\phi}(Z,\Xu | \Yo,\Xo) &= q_{\phi}(Z | \Yo,\Xo)q_{\phi}(\Xu | \Xo) \\
&= \prod_{i=1}^N q_{\phi}(\z | \yo,\xo)q_{\phi}(\xu | \xo).
\end{align*}
We  assume Gaussian variational distributions for the latent variables $\z$, and categorical and Gaussian distributions for discrete and continuous-valued covariates $\xu$, respectively. If the objective is not to perform prediction of unseen data instances but only to impute the missing values, it is possible to replace $q_{\phi}(\xu | \xo)$ with $q_{\phi}(\xu |\yo,\xo)$.

\paragraph{Evidence lower bound with missing covariates} For GP prior VAEs with missing covariates, the ELBO can be written as (see Suppl. Section~\ref{sec:suppl-ELBOderivation} for details) 
\begin{align}
\label{eq:ELBO_2}
&\log p_{\omega}(\Yo | \Xo) \ge \L(\phi,\psi,\theta, \lambda;\Yo, \Xo) \\ 
&\triangleq \E_q[ \log p_{\psi}(\Yo | Z)] \nonumber \\& \quad 
- \mathrm{KL}[q_{\phi}(Z, \Xu | \Yo, \Xo) || p_{\theta,\lambda}(Z,\Xu | \Xo)].\nonumber
\end{align}
For CVAEs, the reconstruction term is replaced with $\E_q[ \log p_{\psi}(\Yo | Z, X)]$. The KL term in eq.~\ref{eq:ELBO_2} above cannot be computed directly since the prior $p_{\theta,\lambda}(Z,\Xu | \Xo)$ represents the joint distribution of the missing auxiliary covariates and the latent variables, which does not have any known closed form expression. For example, in GP prior VAE models, the covariates $\x$ are inputs to the covariance function of the GP that implements the covariance matrix for latent variables, resulting in highly non-trivial joint distributions for $Z$ and $\Xu$. In Suppl.\ Section~\ref{sec:suppl-KLderivation} we show that the KL term can be written in a form that can be computed as:
\begin{align}
    D_{\mathrm{KL}}^1 &\triangleq \mathrm{KL}[q_{\phi}(Z, \Xu | \Yo, \Xo) || p_{\theta,\lambda}(Z,\Xu | \Xo)] \nonumber\\
    &= \E_{q} \left[\mathrm{KL}[q_{\phi}(Z | \Yo, \Xo) || p_{\theta}(Z | \Xu, \Xo)]\right] \label{eq:E-of-KL} \\ & \quad + \mathrm{KL}[q_{\phi}(\Xu |\Xo) || p_{\lambda}(\Xu)],
    \label{eq:E-of-KL2}
\end{align}
where the expectation is w.r.t.\ $q_{\phi}(\Xu | \Xo)$, which can involve both continuous and discrete-valued variables. Note that with our choice of priors and variational families, the KL terms in eqs.~\ref{eq:E-of-KL} and~\ref{eq:E-of-KL2} can be computed analytically. However, the expectation in eq.~\ref{eq:E-of-KL} cannot generally be solved in closed-form. For discrete-valued covariates we compute the expectation exactly by summing over the categorical values, and for continuous-valued covariates we use Monte Carlo sampling.  While eqs.~\ref{eq:ELBO_2}, \ref{eq:E-of-KL} and~\ref{eq:E-of-KL2} serve as the general ELBO for conditional VAEs with missing covariates, it can be further developed for specific models that we will consider next. Note that our choice of independent variational and prior distributions imply that the KL divergence in eq.~\ref{eq:E-of-KL2} factorises across the data samples, $\mathrm{KL}[q_{\phi}(\Xu |\Xo) || p_{\lambda}(\Xu)] = \sum_{i=1}^N \mathrm{KL}[q_{\phi}(\xu_i | \xo_i) || p_{\lambda}(\xu_i)]$, which we will exploit in the following sections.

\paragraph{KL divergence for CVAEs} Because the CVAE model assumes independence across the data samples, the $D_{\mathrm{KL}}^1$ term can be simplified to
\begin{align*}
    D_{\mathrm{KL}}^{\mathrm{CVAE}} &\triangleq \sum_{i=1}^N \Bigl(\E_{q} \left[\mathrm{KL}[q_{\phi}(\z_i | \yo_i, \xo_i) || p_{\theta}(\z_i | \xu_i, \xo_i)]\right] \\ 
    & \quad + \mathrm{KL}[q_{\phi}(\xu_i | \xo_i) || p_{\lambda}(\xu_i)]\Bigl),
\end{align*}
which is directly amenable to mini-batch based SGD. 

\paragraph{KL divergence for GP prior VAEs} 
The multi-output GP prior factorises across the latent dimensions $p_{\theta}(Z|\Xu, \Xo) = \prod_{l=1}^L p_{\theta}(\bar{\z}_l|X)$ (see eq.~\ref{eq:GP-prior-factorizes}) so that the additive nature of the KL divergence for independent distributions can be exploited:
\begin{align}
 D_{\mathrm{KL}}^2 &\triangleq
 \mathrm{KL}[q_{\phi}(Z|\Yo,\Xo)||p_{\theta}(Z|\Xu, \Xo)] \nonumber \\ 
 &= \sum_{l=1}^L \mathrm{KL}[q_{\phi}(\bar{\z}_l|\Yo,\Xo)||p_\theta(\bar{\z}_l|X)] \nonumber \\
 &= \sum_{l=1}^L\mathrm{KL}[\mathcal{N}(\bar{\boldsymbol{\mu}}_l,W_l)||\mathcal{N}(\boldsymbol{0},K_{XX}^{(l)})],\label{eq:KL-l}
\end{align}
where $\bar{\boldsymbol{\mu}}_{l} = ({\mu}_{\phi,l}(\xo_1,\yo_1), \ldots, {\mu}_{\phi,l}(\xo_N,\yo_N))^T$ and $W_l = \mathrm{diag}({\sigma}^2_{\phi,l}(\xo_1,\yo_1), \ldots, {\sigma}^2_{\phi,l}(\xo_N,\y_N))$ are the encoder means and variances, respectively, for the $l^{\text{th}}$ latent dimension.  
Each of the KL divergences above has a closed-form solution, however their exact computation time has cubic scaling, $\mathcal{O}(N^3)$. 

\paragraph{Scalable computation and mini-batching}
We build our scalable computation for eq.~\ref{eq:KL-l} on the fact that any lower bound for the GP prior marginal log-likelihood induces an upper bound for the KL divergence $\mathrm{KL}[\mathcal{N}(\bar{\boldsymbol{\mu}}_l,W_l)||\mathcal{N}(\boldsymbol{0},K_{XX}^{(l)})]$ \citep{ramchandran2021longitudinal}. We use the low-rank inducing point approximation for GPs~\citep{hensman2013gaussian} and introduce $M$ inducing locations $S = (\s_1,\ldots,\s_M)$ in $\X$ and the corresponding inducing function values $\u_l = (f_l(\s_1),\ldots,f_l(\s_M))^T = (u_{l1},\ldots,u_{lM})^T$ for each latent dimension. As proposed in \citet{hensman2013gaussian}, we explicitly keep track of the distribution of inducing values, which is assumed to be Gaussian $\u_l \sim \mathcal{N}(\m_l,H_l)$, where $\m_l$ and $H_l$ are global variational parameters. In the following, we drop the index $l$ for notational simplicity. Following the derivation in~\citet{ramchandran2021longitudinal}, we obtain an upper-bound for the KL divergence $\mathrm{KL}[\mathcal{N}(\bar{\boldsymbol{\mu}},W)||\mathcal{N}(\boldsymbol{0},K_{XX})] \le D_{\mathrm{KL}}^3$ as well as an unbiased, batch-normalised partial sum over a subset of indices, $\mathcal{I} \subset \{1, \ldots, N\}$ of size $|\mathcal{I}| = \hat{N}$ such that $\hat{D}_{\mathrm{KL}}^3 \approx D_{\mathrm{KL}}^3$, where 
\begin{align} 
    \hat{D}_{\mathrm{KL}}^3 &= \frac{1}{2}\frac{N}{\hat{N}} \sum_{i \in \mathcal{I}} \Bigl( \sigma_z^{-2} (K_{\x_iS}K_{SS}^{-1}\m - \bar{\mu}_i)^2 + \sigma_z^{-2} \sigma_i^2 \nonumber \\ &\quad + \sigma_z^{-2} \tilde{K}_{ii} + \sigma_z^{-2} \tr\left( \left(K_{SS}^{-1}HK_{SS}^{-1} \right) \left(K_{S\x_i}K_{\x_iS} \right)\right) \nonumber \\ 
    &\quad - \log \sigma_i^2 \Bigl) + \frac{N}{2} \log \sigma_z^2  - \frac{N}{2} \nonumber \\ &\quad + \mathrm{KL}[\mathcal{N}(\m,H) || \mathcal{N}(\boldsymbol{0},K_{SS})], \label{eq:DKL3hat}
\end{align}
where $\bar{\mu}_i = \bar{\mu}_{\phi}(\xo_i,\yo_i)$ and $\sigma_i^2 = \sigma_\phi^2(\xo_i,\yo_i)$ are the encoder means and variances,
$\tilde{K}_{ii}$ denotes the $i^\mathrm{th}$ diagonal element of $\tilde{K} = K_{XX}-K_{XS}K_{SS}^{-1}K_{SX}$, and $K_{SS}$, $K_{XS} = K_{SX}^T$ and $K_{\x_iS} = K_{S\x_i}^T$ are defined similarly as $K_{XX}$. This scalable upper-bound can be applied for the regression and temporal GP prior VAEs, whereas~\citet{ramchandran2021longitudinal} provide a provably tighter bound for longitudinal GP prior VAEs (see Suppl.\ Section~\ref{sec:suppl-KLUB-L-VAE} for a detailed expression). Note that when using $\hat{D}_{\mathrm{KL}}^3$ from eq.~\ref{eq:DKL3hat} in eq.~\ref{eq:E-of-KL}, we also need to compute the expectation $\E_{q_\phi}[\hat{D}_{\mathrm{KL}}^3]$.

To summarise, the ELBO for GP prior VAE models that marginalises missing covariates and allows efficient optimisation with SGD is obtained from eqs.~\ref{eq:ELBO_2},~\ref{eq:E-of-KL},~\ref{eq:E-of-KL2},~\ref{eq:KL-l}, and~\ref{eq:DKL3hat}.


\paragraph{Learning the parameters of the prior} We make use of an empirical data-driven approach to learn the parameters $\lambda$ of the informative prior $p_{\lambda}(\x)$ instead of assuming an uninformative prior. For continuous covariates, we compute the mean as well as variance of the training data and use that as prior values instead of a standard Gaussian distribution. Similarly, for categorical covariates, we compute the frequency of occurrence for each of the covariates from the training data, which can then be used in the prior distribution instead of a uniform prior. 


\paragraph{Predictive distribution} After training a GP prior VAE model, we want to use it to predict $\y_*$ for previously unseen covariates $\x_*$. Briefly, predictions can be obtained by applying the GP prediction equations in the latent space and then applying the probabilistic decoder that maps from the latent space to the data domain $\Y$. See~\citep{ramchandran2021longitudinal} for details of the predictive equations for the L-VAE model. 

\begin{table*}[!t]
\centering
\caption{NLL values ($\times 10^2$) for predicting $Y$ in the test set  given a partially observed $X$ using CVAE based methods on various versions of the non-temporal rotated digits dataset.}\label{tab:NLL_CVAE}
\resizebox{0.8\textwidth}{!}{%
\begin{tabular}{lc|ccccr}
\hline
\multicolumn{1}{c}{\multirow{2}{*}{\bfseries Method}} & \multicolumn{1}{c|}{\multirow{2}{*}{\bfseries Dataset}} & \multicolumn{5}{c}{\bfseries Missing \%}                                                                  \\
\multicolumn{1}{c}{}                        & \multicolumn{1}{c|}{}                         & \bfseries 5\%               & \bfseries 10\%              & \bfseries 20\%              & \bfseries30\%             & \bfseries 40\%             \\ \hline
Vanilla CVAE                                                         & Dataset 1                                        & $1.6\pm 0.03$                      & $1.7\pm 0.04$                      & $2.4 \pm 0.06$                     & $4.8 \pm 0.08$                     & $4.9 \pm 0.06$                     \\
CVAE with mean imputation                                            & Dataset 1                                                              &$1.5\pm 0.04$  & $1.6\pm 0.03$  & $2.3\pm 0.03$ & $4.6\pm 0.05$  & $4.8\pm 0.04$  \\
CVAE with KNN imputation                                             & Dataset 1                                                              & $1.4\pm 0.03$  & $1.6\pm 0.04$  & $2.3\pm 0.04$  & $4.5\pm 0.03$  & $4.6\pm 0.03$  \\
\rowcolor{Gray}
\textbf{CVAE with our method}                                        & Dataset 1                                                              & $\boldsymbol{1.1\pm 0.02}$                      & $\boldsymbol{1.4\pm 0.03}$                      & $\boldsymbol{2.1\pm 0.05}$                      & $\boldsymbol{4.1\pm 0.03}$                      & $\boldsymbol{4.2 \pm 0.02}$                     \\
CVAE with oracle                                                               & Dataset 1                                         & $0.9\pm 0.01$                      & $1.1\pm 0.01$                      & $1.5 \pm 0.02$                     & $3.1 \pm 0.02$                     & $3.8 \pm 0.02$                     \\ \hline
Vanilla CVAE                                                         & Dataset 2                                                              & $0.43\pm 0.03$                     & $0.81\pm 0.02$                     & $0.65\pm 0.02$                     & $0.71\pm 0.04$                     & $0.81\pm 0.04$                     \\
CVAE with mean imputation                                            & Dataset 2                                                              & $0.41\pm 0.03$ & $0.75\pm 0.03$ & $0.64\pm 0.03$ & $0.69\pm 0.05$ & $0.79\pm 0.04$ \\
CVAE with KNN imputation                                             & Dataset 2                                         & $0.39\pm 0.05$                     & $0.71\pm 0.04$                     & $0.62\pm 0.05$                     & $0.67\pm 0.03$                     & $0.78\pm 0.03$                     \\
\rowcolor{Gray}
\textbf{CVAE with our method}                                                 & Dataset 2                                                              & $\boldsymbol{0.32\pm 0.02}$                     & $\boldsymbol{0.63\pm 0.03}$                     & $\boldsymbol{0.52\pm 0.04}$                     & $\boldsymbol{0.64\pm 0.05}$                     & $\boldsymbol{0.72\pm 0.05}$                     \\
CVAE with oracle                                                               & Dataset 2                                         & $0.1 \pm 0.01$                     & $0.44\pm 0.01$                     & $0.48\pm 0.02$                     & $0.52\pm 0.03$                     & $0.53\pm 0.02$                     \\ \hline
\end{tabular}}
\end{table*}

\begin{table*}[!t]
\centering
\caption{NLL values ($\times 10^2$) for predicting $Y$ in the test set  given a partially observed $X$ using GP prior VAE (L-VAE) based methods on various versions of the non-temporal and temporal rotated digits dataset.}\label{tab:NLL_LVAE}
\resizebox{0.8\textwidth}{!}{%
\begin{tabular}{lc|ccccr}
\hline
\multicolumn{1}{c}{\multirow{2}{*}{\bfseries Method}} & \multirow{2}{*}{\bfseries Dataset} & \multicolumn{5}{c}{\bfseries Missing \%}                                                                                                     \\
\multicolumn{1}{c}{}                                                 &                                                   & \bfseries 5\% & \bfseries 10\% & \bfseries 20\% & \bfseries30\% & \bfseries 40\% \\ \hline
Regression L-VAE with mean impute                                               & Dataset 1                                & $0.72\pm 0.007$              & $0.74\pm 0.005$               & $1.4\pm 0.007$                & $2.8\pm 0.01$                & $3.6\pm 0.01$                 \\
Regression L-VAE with KNN impute                                                & Dataset 1                               & $0.62\pm 0.005$              & $0.71\pm 0.006$               & $1.1\pm 0.008$                & $2.3\pm 0.008$               & $3.1\pm 0.01$                 \\
\rowcolor{Gray}
\textbf{Regression L-VAE with our method}                                                & Dataset 1                               & $\boldsymbol{0.58\pm 0.004}$              & $\boldsymbol{0.6\pm 0.004}$                & $\boldsymbol{0.81\pm 0.004}$               & $\boldsymbol{1.1\pm 0.004}$               & $\boldsymbol{2.3\pm 0.009}$                \\
Regression L-VAE with oracle                                                    & Dataset 1                                & $0.55\pm 0.003$              & $0.58\pm 0.002$               & $0.74\pm 0.004$               & $0.82\pm 0.002$              & $1.8\pm 0.006$                \\  \hline
Regression L-VAE with mean impute                                               & Dataset 2                       & $0.21\pm 0.02$               & $0.25\pm 0.009$               & $0.36\pm 0.01$                & $0.48\pm 0.03$               & $0.51\pm 0.02$                \\
Regression L-VAE with KNN impute                                                & Dataset 2                          & $0.18\pm 0.02$               & $0.22\pm 0.01$                & $0.32\pm 0.02$                & $0.41\pm 0.03$               & $0.56\pm 0.03$                \\
\rowcolor{Gray}
\textbf{Regression L-VAE with our method}                                                & Dataset 2                         & $\boldsymbol{0.14\pm 0.003}$              & $\boldsymbol{0.19\pm 0.002}$               & $\boldsymbol{0.26\pm 0.004}$               & $\boldsymbol{0.29\pm 0.05}$               & $\boldsymbol{0.39\pm 0.008}$               \\ 
Regression L-VAE with oracle                                                    & Dataset 2                          & $0.12\pm 0.003$              & $0.18\pm 0.003$               & $0.21\pm 0.004$               & $0.26\pm 0.006$              & $0.36\pm 0.006$               \\ \hline
Temporal L-VAE with mean impute                                               & Dataset 3                                    & $0.19\pm 0.008$              & $0.27\pm 0.008$               & $0.39\pm 0.02$                & $0.42\pm 0.01$               & $0.56\pm 0.03$                \\
Temporal L-VAE with KNN impute                                                & Dataset 3                                   & $0.17\pm 0.005$              & $0.25\pm 0.009$               & $0.35\pm 0.01$                & $0.47\pm 0.02$               & $0.58\pm 0.03$                \\
\rowcolor{Gray}
\textbf{Temporal L-VAE with our method}                                                & Dataset 3                                    & $\boldsymbol{0.12\pm 0.002}$              & $\boldsymbol{0.19\pm 0.003}$               & $\boldsymbol{0.25\pm 0.006}$               & $\boldsymbol{0.38\pm 0.01}$               & $\boldsymbol{0.44\pm 0.02}$                \\ 
Temporal L-VAE with oracle                                                    & Dataset 3                                    & $0.11\pm 0.001$              & $0.16\pm 0.003$               & $0.21\pm 0.007$               & $0.28\pm 0.008$              & $0.37\pm 0.01$                \\
\hline
\end{tabular}}
\end{table*}

\begin{table*}[!t]
\centering
\caption{MSE values for covariate imputation on various versions of the rotated digits dataset.}\label{tab:MSE_labels}
\resizebox{0.8\textwidth}{!}{%
\begin{tabular}{lc|ccccr}
\hline
\multicolumn{1}{c}{\multirow{2}{*}{\bfseries Method}} & \multicolumn{1}{c|}{\multirow{2}{*}{\bfseries Dataset}} & \multicolumn{5}{c}{\bfseries Missing \%}                                                                         \\
\multicolumn{1}{c}{}                        & \multicolumn{1}{c|}{}                         & \bfseries 5\%             & \bfseries 10\%            & \bfseries 20\%                           & \bfseries 30\%            & \bfseries 40\%            \\ \hline
Mean impute                      & Dataset 1                              & $0.225$         & $0.45$          & $0.52$                         & $0.87$          & $1.025$         \\
KNN impute                       & Dataset 1                            & $0.175$         & $0.425$         & $0.51$                         & $0.575$         & $0.823$         \\
\rowcolor{Gray}
\textbf{C-VAE with our method}                       & Dataset 1                              & $\boldsymbol{0.19\pm 0.08}$  & $\boldsymbol{0.23\pm 0.05}$  & $\boldsymbol{0.3\pm0.06}$                   & $\boldsymbol{0.45\pm 0.08}$  & $\boldsymbol{0.61\pm 0.05}$  \\
\rowcolor{Gray}
\textbf{Regression L-VAE with our method}                       & Dataset 1                              & $\boldsymbol{0.075\pm 0.05}$ & $\boldsymbol{0.1\pm 0.06}$   & $\boldsymbol{0.225\pm0.05}$                 & $\boldsymbol{0.28\pm 0.06}$  & $\boldsymbol{0.39\pm 0.06}$   \\ \hline
Mean impute                      & Dataset 2                        & $0.15$          & $0.425$         & $0.55$                         & $0.925$         & $1.05$          \\
KNN impute                       & Dataset 2                             & $0.275$         & $0.31$          & $0.45$                         & $0.825$         & $0.975$         \\
\rowcolor{Gray}
\textbf{C-VAE with our method}                       & Dataset 2                      & $\boldsymbol{0.1\pm 0.01}$   & $\boldsymbol{0.275\pm 0.04}$ & $\boldsymbol{0.41\pm 0.07}$                 & $\boldsymbol{0.71\pm 0.06}$  & $\boldsymbol{0.725\pm 0.06}$ \\
\rowcolor{Gray}
\textbf{Regression L-VAE with our method}                       & Dataset 2                      & $\boldsymbol{0.05\pm 0.04}$  & $\boldsymbol{0.125\pm 0.04}$ & $\boldsymbol{0.26\pm 0.06}$                 & $\boldsymbol{0.48\pm 0.05}$  & $\boldsymbol{0.61\pm 0.04}$  \\ \hline
Mean impute                      & Dataset 3                                & $0.75$          & $0.77$          & $0.95$                         & $0.97$          & $1.25$          \\
KNN impute                       & Dataset 3                                & $0.71$          & $0.74$          & $1.025$                        & $1.012$         & $1.35$          \\
\rowcolor{Gray}
\textbf{Temporal L-VAE with our method}                       & Dataset 3                                & $\boldsymbol{0.21\pm 0.02}$  & $\boldsymbol{0.22\pm 0.04}$  & $\boldsymbol{0.29\pm 0.06}$                 & $\boldsymbol{0.35\pm 0.04}$  & $\boldsymbol{0.68\pm 0.06}$ \\
\hline
\end{tabular}}
\end{table*}
\section{Experiments}
\label{sec:experiments}
We quantify the efficacy of our method by performing the imputation of missing values as well as predicting unseen instances of $Y$ for a synthetic dataset as well as a real-world longitudinal clinical trial  dataset. We demonstrate the improvements offered by our method on different conditional generative models, namely CVAE and GP prior VAE models. The performance of our method is compared against various alternative approaches to handle missing auxiliary covariate values. To provide fair comparisons, we have used similar encoder and decoder neural network architectures when performing comparisons across different methods. 

\paragraph{Rotated digits dataset} The dataset comprises of several measurements of a digit from the MNIST dataset \citep{lecun1998gradient}. Each measurement is manipulated by a rotation about the centre of the digit, a translation (or shift) along the diagonal, and intensity of the digit (or contrast). The individual pixel values of the manipulated digit forms the data $Y$. The rotation, diagonal shift, and image contrast form the auxiliary covariate information $X$. We experimented with various variants of the dataset. For \textbf{non-temporal rotated digits dataset} we consider two variants. The \textbf{independent covariates} dataset (`Dataset 1' in the experiments) is obtained by randomly sampling the covariate values from a Gaussian distribution, where the covariate dimensions are independent of each other. The \textbf{dependent covariates} dataset (`Dataset 2' in the experiments) is obtained by sampling the covariate values from a distribution, where the covariate dimensions are correlated and have a non-trivial correlation structure (that does not follow any specific parametric form). The \textbf{temporal rotated digits dataset} (`Dataset 3' in the experiments) is a temporal variant in which the auxiliary covariate information are temporally correlated through a time covariate which is sampled between $[t_{\min},t_{\max}]$ and included in $X$. For each observation time point $t$, we randomly draw the other covariates (i.e., rotation, shift and contrast) via a mapping that involves a non-linear transformation and additive Gaussian-distributed variation, resulting in a non-trivial distribution that does not follow any parametric form. Details of the data simulation and visualisations of the covariate distributions as well as transformed images can be found in Suppl.\ Section~\ref{seq:suppl-data-generation}. The training set comprised of $4000$ observations, the validation set used for early-stopping comprised of $400$ observations, and the test set comprised of $400$ observations.

\paragraph{Clinical trial dataset} The longitudinal dataset that we used is a randomised clinical drug trial for the treatment of Prostate cancer obtained from $371$ patients who were observed over a period of $2$ years at most. The dataset is available on Project Datasphere's open-access platform \citep{green2015project}. In this dataset, about $25.01\%$ of the lab measurements were observed. After filtering out the patients with insufficient number of samples, we were left with $184$ patients. More information on the data pre-processing can be found in Suppl.\ Section~\ref{sec:suppl-clinical-data-description}. The training set comprised of $144$ patients with a total of $979$ observations (each patient has at least $5$ observations). The validation set used for early stopping comprised of $20$ patients with a total of $141$ observations, and the test set comprised of $20$ patients with a total of $167$ observations. Each patient has $28$ lab measurements as well as vital signs that are partially observed ($Y$) and $23$ patient-specific auxiliary covariates ($X$). The patient-specific auxiliary covariates are also partially observed, and include $10$ possible adverse events that may be experienced by a patients and $10$ concomitant medications that may be used by the patient. In addition to this, the auxiliary covariates include the time elapsed from the start of the study, the patient's age as well as sex. 

\subsection{Experiments with CVAEs}
\label{subsec:experiments_cvae}
We demonstrate the benefits of using our proposed method in CVAEs by comparing the performance of various alternative models in predicting $Y$ and the missing auxiliary covariates $X$ of the test dataset. In this experiment, we make use of the \textbf{non-temporal rotated digits} dataset (Dataset 1 and Dataset 2). Moreover, we vary the amount of missing data in $X$ and $Y$ of the training, test, and validation set. 

We compare a vanilla CVAE in which $\xu$ and $\yu$ are replaced by $0$ and a CVAE in which the missing values are marginalised using our method. Moreover, we also compare the performance of using a CVAE in which the missing values in $x$ are imputed using mean imputation and k-Nearest Neighbours (k-NN) imputation. As a baseline, we also include the scenario in which $\x$ is fully observed and contains no missing values (the oracle model). Table~\ref{tab:NLL_CVAE} shows the results of the models in terms of negative log likelihood (NLL) for predicting $Y$ of the test set given a partially observed $X$ of the test set. 
It can be seen that the performance of CVAE improves by making use of our proposed marginalisation method. Moreover, our method's performance is generally close to that of the oracle method (CVAE with oracle) that makes use of fully observed covariates.


\subsection{Experiments with GP prior VAEs}
\label{subsec:experiments_tempGPVAE}
We demonstrate the improvement afforded by our method to GP prior VAEs using the L-VAE model. 
We compare the performance of L-VAE enhanced with our method, with the original L-VAE where the missing values in $\x$ are imputed using mean imputation and k-Nearest Neighbours (k-NN) imputation. We also compare our method to the baseline scenario in which $\x$ is fully observed 
(the oracle model). We compare the performance of the models in terms of negative log likelihood (NLL) for predicting $Y$ of the test set given a partially observed $X$ of the test set in table~\ref{tab:NLL_LVAE}. The experiments show that GP prior VAEs with our marginalisation method perform better at predicting the unseen test dataset. The MSE values of the imputed auxiliary covariates $X$ of the test set (using the mean of $q_{\phi}(\xu | \xo)$) can be seen in table \ref{tab:MSE_labels}. The lower MSE values in table \ref{tab:MSE_labels} obtained by our method show that we are able to produce better estimates of the missing auxiliary covariates. 

We also demonstrate the performance of our method on a variant of the \textbf{temporal rotated digits} dataset that contains also a discrete covariate in Suppl. tables \ref{tab:MSE_het_cVAE} to \ref{tab:MSE_label_disc_cov}.

\begin{figure}[t]
\centering
  \includegraphics[width=0.8\linewidth]{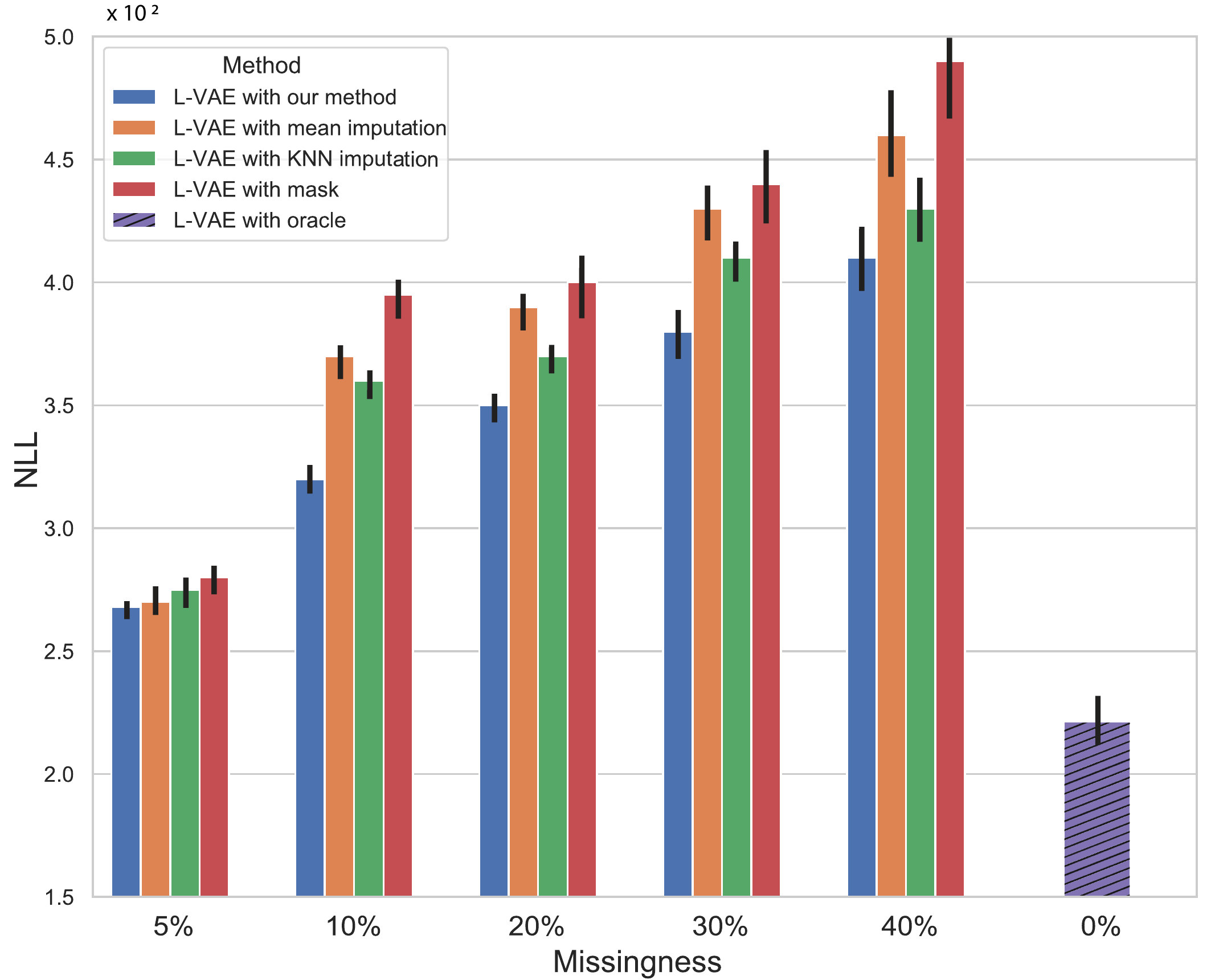}
  \caption{NLL values for the test set of the clinical trial dataset.}
  \label{fig:nll_clincial_trial}
\end{figure}
\subsection{Experiments with Longitudinal GP prior VAEs}
\label{subsec:longGPVAE}
We demonstrate the efficacy of our method on longitudinal data using a real-world \textbf{clinical trial dataset} described above. In particular we compare the performance of our approach with alternative approaches by comparing the predictive performance on the lab measurements of an unseen test set. \citet{ramchandran2021longitudinal} have shown that longitudinal GP prior VAEs are well suited to longitudinal datasets. 
Therefore, we shall use the L-VAE model with the additive covariance function as described in Section~\ref{subsec:gpvae} as the base model to which we apply our approach.

We experimented with different amounts of missing values (5\%, 10\%, 20\%, 30\%, and 40\%) artificially introduced completely at random into the auxiliary covariates of the training set, validation set, and test set. We measured the model's ability to generate the test data given just the partially observed auxiliary covariates of the test set. 
Our method estimates the missing auxiliary covariates using the proposed amortised variational inference and then generates the data $Y$ of the test set by building upon L-VAE. Fig.~\ref{fig:nll_clincial_trial} compares the NLL obtained by L-VAE enhanced with our method, with other methods for auxiliary covariate imputation. The auxiliary covariates in L-VAE with oracle are fully-observed and hence, demonstrates the lowest possible NLL. The lower NLL obtained by using our method with L-VAE shows that we are able to achieve better prediction performance even for longitudinal datasets. The MSE of the imputed auxiliary covariates $X$ of the test set can be seen in Suppl. table \ref{tab:MSE_label_clinical}.
\label{subsec:experiments_longGPVAE}

\section{Discussion}
\label{sec:discussion}
In this paper, we introduced a novel method to improve the performance of conditional VAEs on datasets in which the auxiliary covariates are partially observed. We achieve this by assigning a prior distribution for the missing covariates and estimating their posterior distribution using amortised variational inference. The method that we proposed is applicable to a variety of conditional VAE models, including but not limited to, CVAEs and GP prior VAEs.
Furthermore, we derive computationally efficient evidence lower bounds that make use of mini-batching for CVAE and GP prior based methods. The efficacy of our proposed method was demonstrated on synthetic toy datasets as well as a real-world clinical trial dataset. Our experiments focused on the benefits of simultaneously estimating the missing auxiliary covariates along with the missing observations in conditional VAE models. Given the wide applicability of this work, we believe that our method would be important in the development of conditional VAE models.

%

\begin{acknowledgements} 
We would like to acknowledge the computational resources provided by Aalto Science-IT, Finland. We would also like to thank Manuel Haussmann for the helpful discussions and comments.
\end{acknowledgements}

\bibliography{references}

\newpage
\onecolumn

\appendix

\section{Supplementary methods}

\subsection{Evidence lower bound for conditional VAEs with missing covariates}
\label{sec:suppl-ELBOderivation}

Variational inference seeks to minimise the KL divergence from the variational approximation to the true posterior, which is known to correspond to the maximisation of the ELBO. Although the derivation follows closely the standard derivation, here we show in detail how the ELBO is obtained for the general class of conditional VAEs (including CVAEs and GP prior VAEs) in the case of missing values in data $Y$ and auxiliary covariates $X$. We start with the GP prior VAE model and note that using the conditional probability, the true posterior can be written as
\begin{align}
    p_{\omega}(Z,\Xu | \Yo, \Xo) &= \frac{p_{\psi}(\Yo | Z,\Xu, \Xo) p_{\theta,\lambda}(Z,\Xu | \Xo)}{p_{\omega}(\Yo | \Xo)} \label{eq:supp-condprob-cvae} \\ 
    &= \frac{p_{\psi}(\Yo | Z) p_{\theta,\lambda}(Z,\Xu | \Xo)}{p_{\omega}(\Yo | \Xo)}. \label{eq:supp-condprob}
\end{align}
The KL that we want to minimise can then be written using eq.~\ref{eq:supp-condprob} as
\begin{align}
    \mathrm{KL}[q_{\phi}(Z, \Xu | \Yo, \Xo) || p_{\omega}(Z, \Xu | \Yo, \Xo)] &= \E_q\left[\log \frac{q_{\phi}(Z, \Xu | \Yo, \Xo)}{p_{\omega}(Z, \Xu | \Yo, \Xo)}\right] \nonumber \\
    &= \E_q\left[\log \frac{q_{\phi}(Z, \Xu | \Yo, \Xo)p_{\omega}(\Yo | \Xo)}{p_{\psi}(\Yo | Z) p_{\theta,\lambda}(Z,\Xu | \Xo)}\right] \nonumber \\
    &= \E_q[\log q_{\phi}(Z, \Xu | \Yo, \Xo)] + \E_q[\log p_{\omega}(\Yo | \Xo)] \nonumber \\ 
    & \quad - \E_q[ \log p_{\psi}(\Yo | Z)] - \E_q[\log p_{\theta,\lambda}(Z,\Xu | \Xo)]. \nonumber 
\end{align}
Note that the expectations above are w.r.t.\ $Z$ or both $Z$ and $\Xu$, and that expectations w.r.t.\ $\Xu$ can contain both continuous and discrete-valued variables. Noting that $\E_q[\log p_{\omega}(\Yo | \Xo)] = \log p_{\omega}(\Yo | \Xo)$ and that the KL is always non-negative, by rearranging we obtain the lower bound
\begin{align}
\log p_{\omega}(\Yo | \Xo) & \ge \E_q[ \log p_{\psi}(\Yo | Z)] - \E_q[\log q_{\phi}(Z, \Xu | \Yo, \Xo)] + \E_q[\log p_{\theta,\lambda}(Z,\Xu | \Xo)] \nonumber \\
& = \E_q[ \log p_{\psi}(\Yo | Z)] - \mathrm{KL}[q_{\phi}(Z, \Xu | \Yo, \Xo) || p_{\theta,\lambda}(Z,\Xu | \Xo)] \label{eq:supp-elbokl}\\
& \triangleq \L(\phi,\psi,\theta, \lambda;\Yo, \Xo). \nonumber 
\end{align}

The above ELBO is specific for the GP prior VAE model. The derivation for the CVAE model is otherwise the same except that in eq.~\ref{eq:supp-condprob-cvae} the conditional probability $p_{\psi}(\Yo | Z,\Xu, \Xo) = p_{\psi}(\Yo | Z,X)$ cannot be simplified further as $\Yo$ depends also on $X$. Following the same steps as above the ELBO for CVAE can be written as
\begin{equation*}
\L(\phi,\psi,\theta, \lambda;\Yo, \Xo) = \E_q[ \log p_{\psi}(\Yo | Z, X)] - \mathrm{KL}[q_{\phi}(Z, \Xu | \Yo, \Xo) || p_{\theta,\lambda}(Z,\Xu | \Xo)]. 
\end{equation*}

\subsection{Derivation of the KL divergence}
\label{sec:suppl-KLderivation}

The KL term in eq.~\ref{eq:ELBO_2} (or in eq.~\ref{eq:supp-elbokl}) can be written as follows
\begin{eqnarray*}
\mathrm{KL} &=& \mathrm{KL}[q_{\phi}(Z,\Xu | \Xo, \Yo) || p_{\theta,\lambda}(Z,\Xu | \Xo)]\\ 
&=& \E_{q_{\phi}(Z,\Xu | \Xo, \Yo)} \left[ \log \frac{q_{\phi}(Z,\Xu | \Xo, \Yo)}{p_{\theta,\lambda}(Z,\Xu | \Xo)} \right] \\
&=& \E_{q_{\phi}(Z | \Yo, \Xo)q_{\phi}(\Xu | \Xo)} \left[ \log \frac{q_{\phi}(Z | \Yo, \Xo)q_{\phi}(\Xu | \Xo)}{p_{\theta}(Z | \Xu, \Xo)p_{\lambda}(\Xu | \Xo)} \right] \\
&=& \E_{q_{\phi}(Z | \Yo, \Xo)q_{\phi}(\Xu | \Xo)} \left[ \log \frac{q_{\phi}(Z | \Yo, \Xo)}{p_{\theta}(Z | \Xu, \Xo)} \right] + \E_{q_{\phi}(Z | \Yo, \Xo)q_{\phi}(\Xu | \Xo)} \left[ \log \frac{q_{\phi}(\Xu | \Xo)}{p_{\lambda}(\Xu | \Xo)} \right] \\
&=& \E_{q_{\phi}(\Xu |  \Xo)} \left[ \E_{q_{\phi}(Z | \Yo, \Xo)} \left[ \log \frac{q_{\phi}(Z | \Yo, \Xo)}{p_{\theta}(Z | \Xu, \Xo)} \right] \right] + \E_{q_{\phi}(\Xu | \Xo)} \left[ \log \frac{q_{\phi}(\Xu |  \Xo)}{p_{\lambda}(\Xu | \Xo)} \right] \\
&=& \E_{q_{\phi}(\Xu | \Xo)} [ \mathrm{KL}[q_{\phi}(Z | \Yo, \Xo) || p_{\theta}(Z | \Xu, \Xo)] ] + \mathrm{KL}[q_{\phi}(\Xu |  \Xo) || p_{\lambda}(\Xu | \Xo)]. 
\end{eqnarray*}
Note again that the expectations w.r.t.\ $\Xu$ can involve both continuous and discrete-valued covariates. For our choice of independent prior for missing covariates we have $p_{\lambda}(\Xu | \Xo) = p_{\lambda}(\Xu)$, 

\subsection{Scalable mini-batch compatible KL upper bound for the L-VAE model}
\label{sec:suppl-KLUB-L-VAE}
In a longitudinal setting, we need to separate the additive component that corresponds to the interaction between instances and time (or age) from the other additive components, the covariance matrix has the following general form $\Sigma = K_{XX}^{(A)} + \hat{\Sigma}$, where $\hat{\Sigma} = \text{diag}( \hat{\Sigma}_1, \dots, \hat{\Sigma}_P )$, $\hat{\Sigma}_p = K_{X_pX_p}^{(R)} + \sigma_z^2 I_{n_p}$, and $K_{XX}^{(A)} = \sum_{r=1}^{R-1} K_{XX}^{(r)}$ contains all the other $R-1$ components.

Assume that $\mathcal{I}_{pi}$ is the index of the $i$\textsuperscript{th} sample for the $p$\textsuperscript{th} patient and $\hat{\boldsymbol{\mu}}_p = [\bar{\mu}_{\mathcal{I}_{p1}},\ldots,\bar{\mu}_{\mathcal{I}_{pn_p}}]^T$ is a sub-vector of $\bar{\boldsymbol{\mu}}$ that corresponds to the $p$\textsuperscript{th} patient. Therefore, we have a batch-normalised partial sum over a subset of indices $\mathcal{P} \subset \{1, \ldots, P\}$ of size $|\mathcal{P}| = \hat{P}$:
\begin{align}
    \hat{D}_{\mathrm{KL}}^4 &= \frac{1}{2}\frac{P}{\hat{P}} \sum_{p \in \mathcal{P}} \Biggl( (K_{X_pS}^{(A)}{K_{SS}^{(A)}}^{-1}\m - \hat{\boldsymbol{\mu}}_p)^T \hat{\Sigma}_p^{-1} (K_{X_pS}^{(A)}{K_{SS}^{(A)}}^{-1}\m - \hat{\boldsymbol{\mu}}_p) + \sum_{i=1}^{n_p}{(\hat{\Sigma}_p^{-1})}_{ii} \sigma_\phi^2(\y_{\mathcal{I}_{pi}}) + \log |\hat{\Sigma}_p| \nonumber\\ &\quad + \text{tr} \left( \hat{\Sigma}_p^{-1} \tilde{K}^{(A)}_{X_pX_p} \right) + 
    \tr\left( \left({K_{SS}^{(A)}}^{-1}H{K_{SS}^{(A)}}^{-1} \right) \left(K_{SX_p}^{(A)}\hat{\Sigma}_p^{-1}K_{X_pS}^{(A)} \right)\right) - \sum_{i=1}^{n_p}\log \sigma_\phi^2(\y_{\mathcal{I}_{pi}}) \Biggl)\nonumber\\ &\quad - \frac{N}{2} + \mathrm{KL}[\mathcal{N}(\m,H) || \mathcal{N}(\boldsymbol{0},K_{SS}^{(A)})], 
\end{align}
where $\tilde{K}^{(A)}_{X_pX_p} =K_{X_pX_p}^{(A)}-K_{X_pS}^{(A)}{K_{SS}^{(A)}}^{-1}K_{SX_p}^{(A)}$. This is an unbiased estimate of the KL divergence upper bound $E_{\mathcal{P} \sim \mathfrak{S}\{1, \ldots, P\}} (\hat{D}_{\mathrm{KL}}^4) = D_{\mathrm{KL}}^4 \geq D_{\mathrm{KL}}(\mathcal{N}(\bar{\boldsymbol{\mu}},W)||\mathcal{N}(\boldsymbol{0},\Sigma))$. This property enables us to use the mini-batching technique for a more precise approximate computation of the KL divergence term of L-VAE and its gradients, by approximately splitting equal number of patients to each batch. For a more detailed derivation, please refer to \citet{ramchandran2021longitudinal}.

\section{Generation of the rotated digit dataset}
\label{seq:suppl-data-generation}
\begin{figure}[!h]
\centering
\begin{subfigure}{.33\textwidth}
  \centering
  \includegraphics[width=\linewidth]{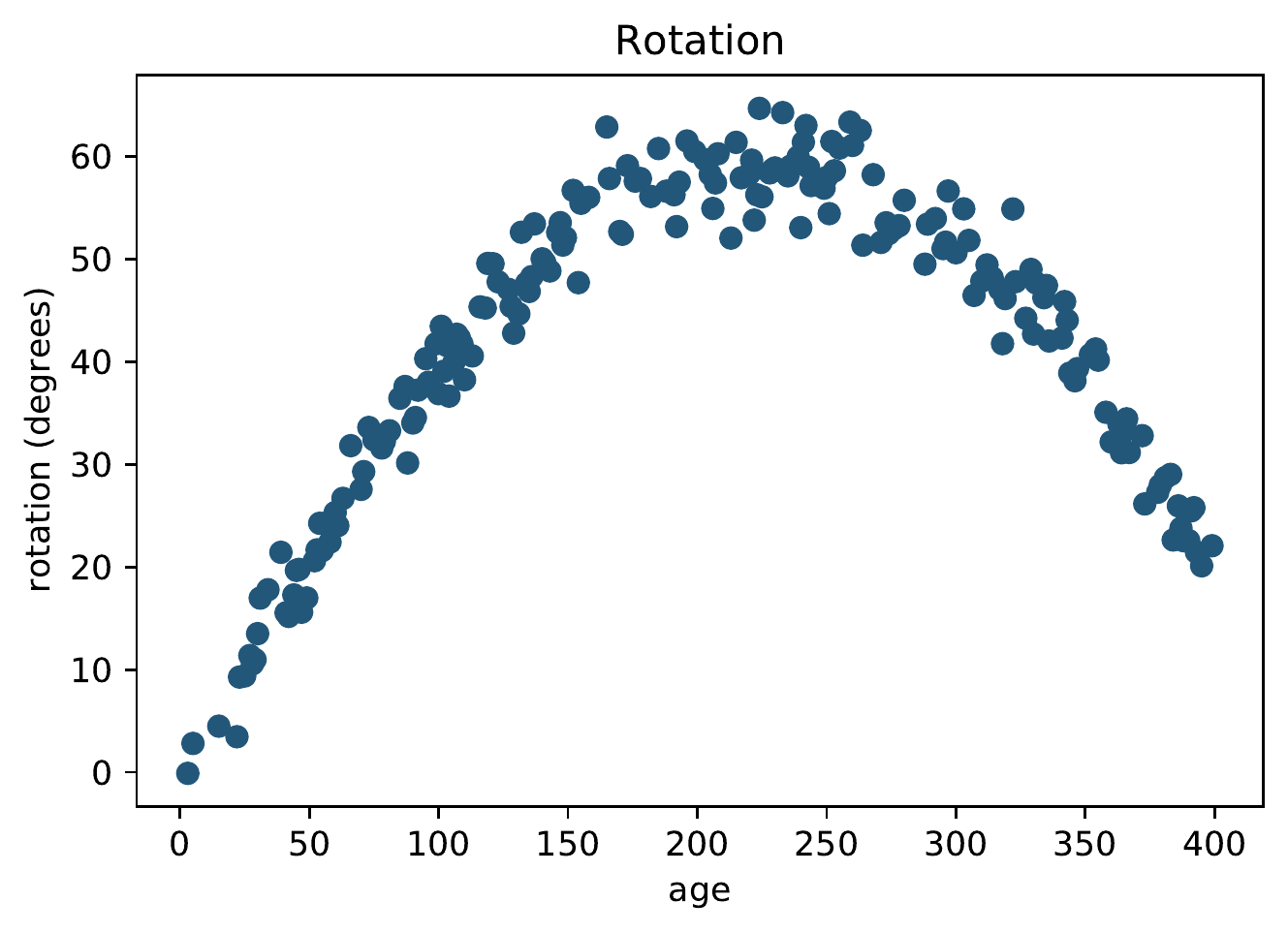}
  \label{fig:train_label_rotation_noise}
\end{subfigure}%
\begin{subfigure}{.33\textwidth}
  \centering
  \includegraphics[width=\linewidth]{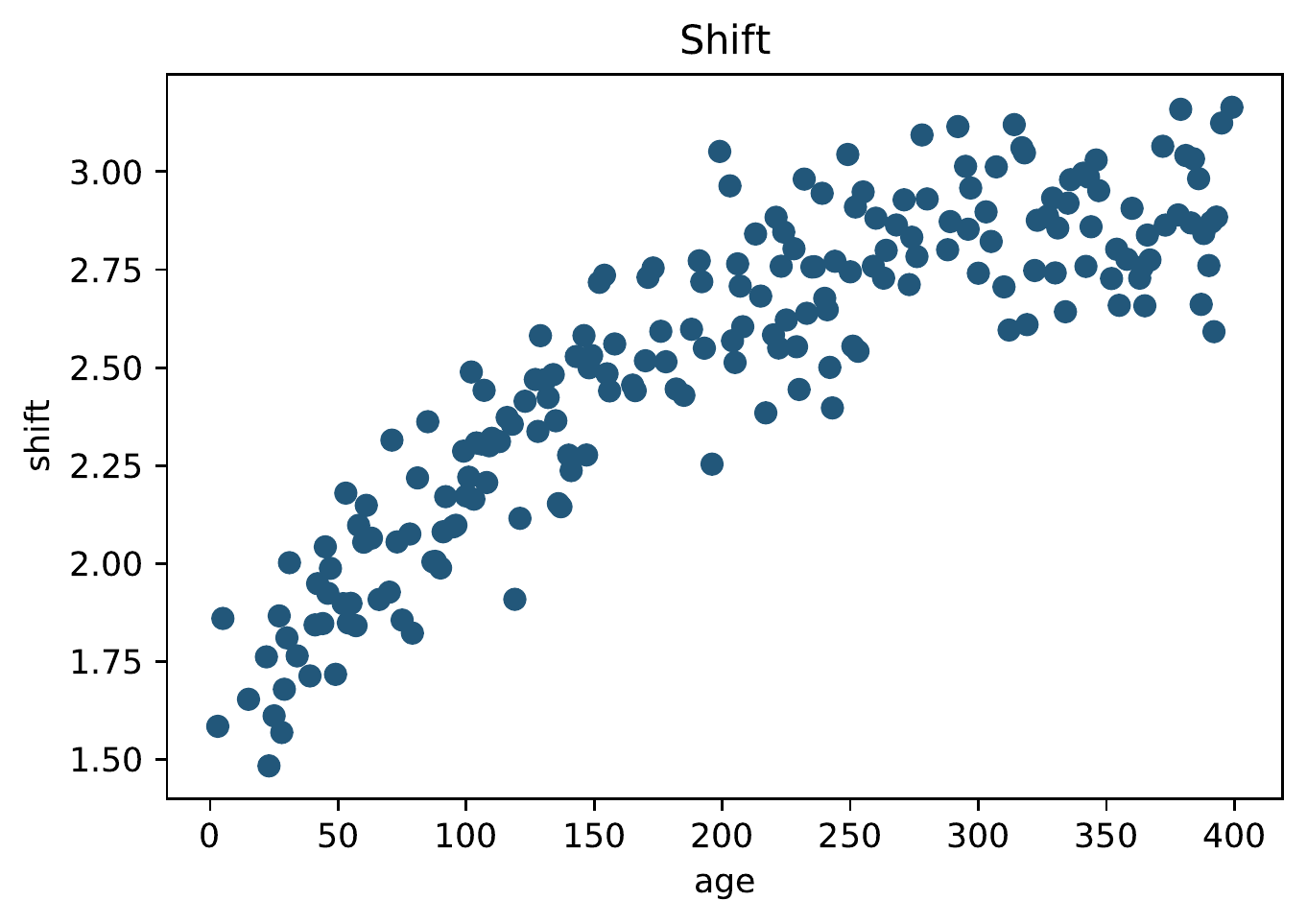}
  \label{fig:train_label_rotation_shift}
\end{subfigure}%
\begin{subfigure}{.33\textwidth}
  \centering
  \includegraphics[width=\linewidth]{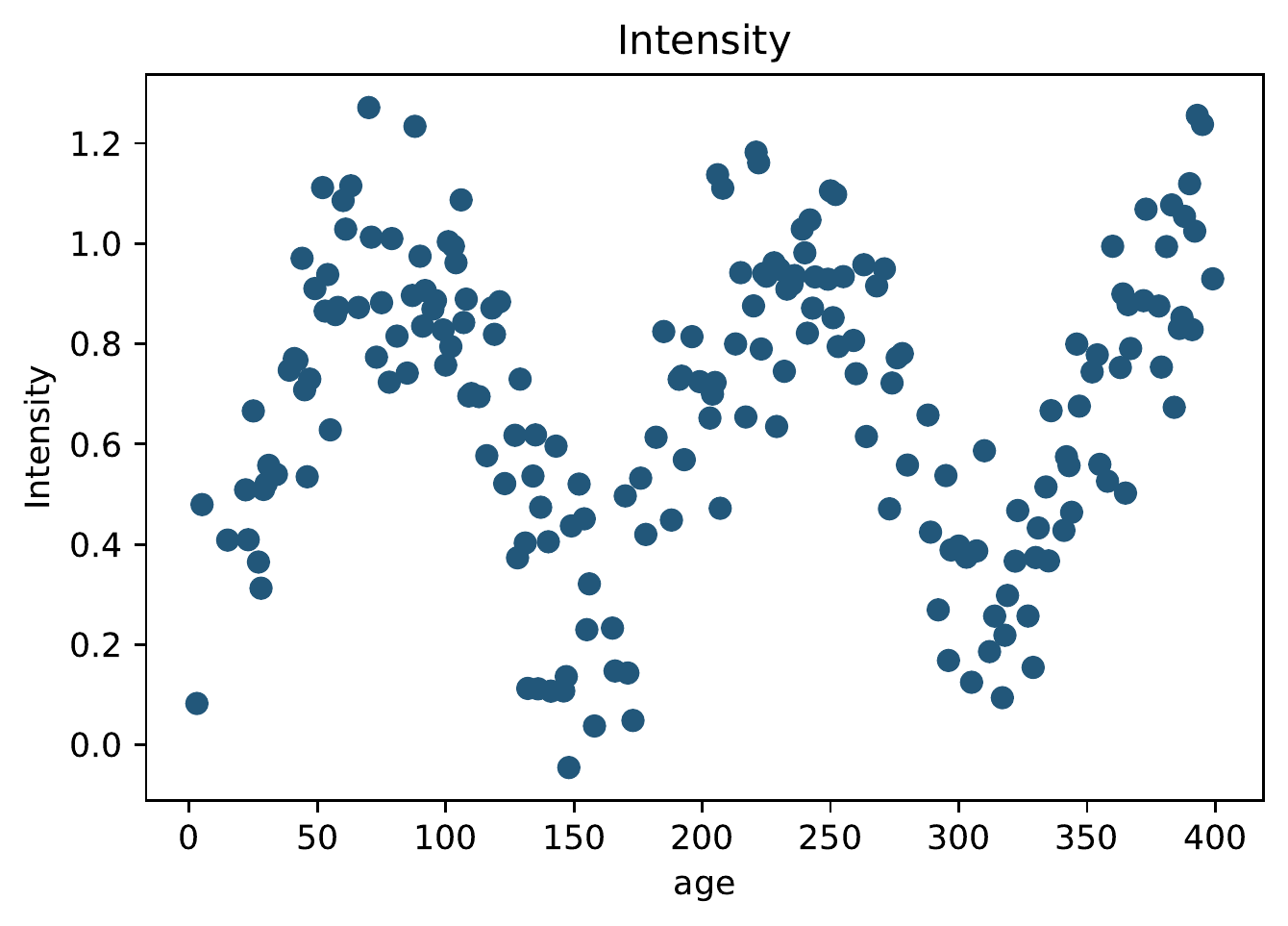}
  \label{fig:train_label_rotation_intensity}
\end{subfigure}
\caption{Visualisation of the auxiliary covariates $X$.}
\label{fig:vis_aux}
\end{figure}

\begin{figure}[h]
\centering
  \includegraphics[width=0.6\linewidth]{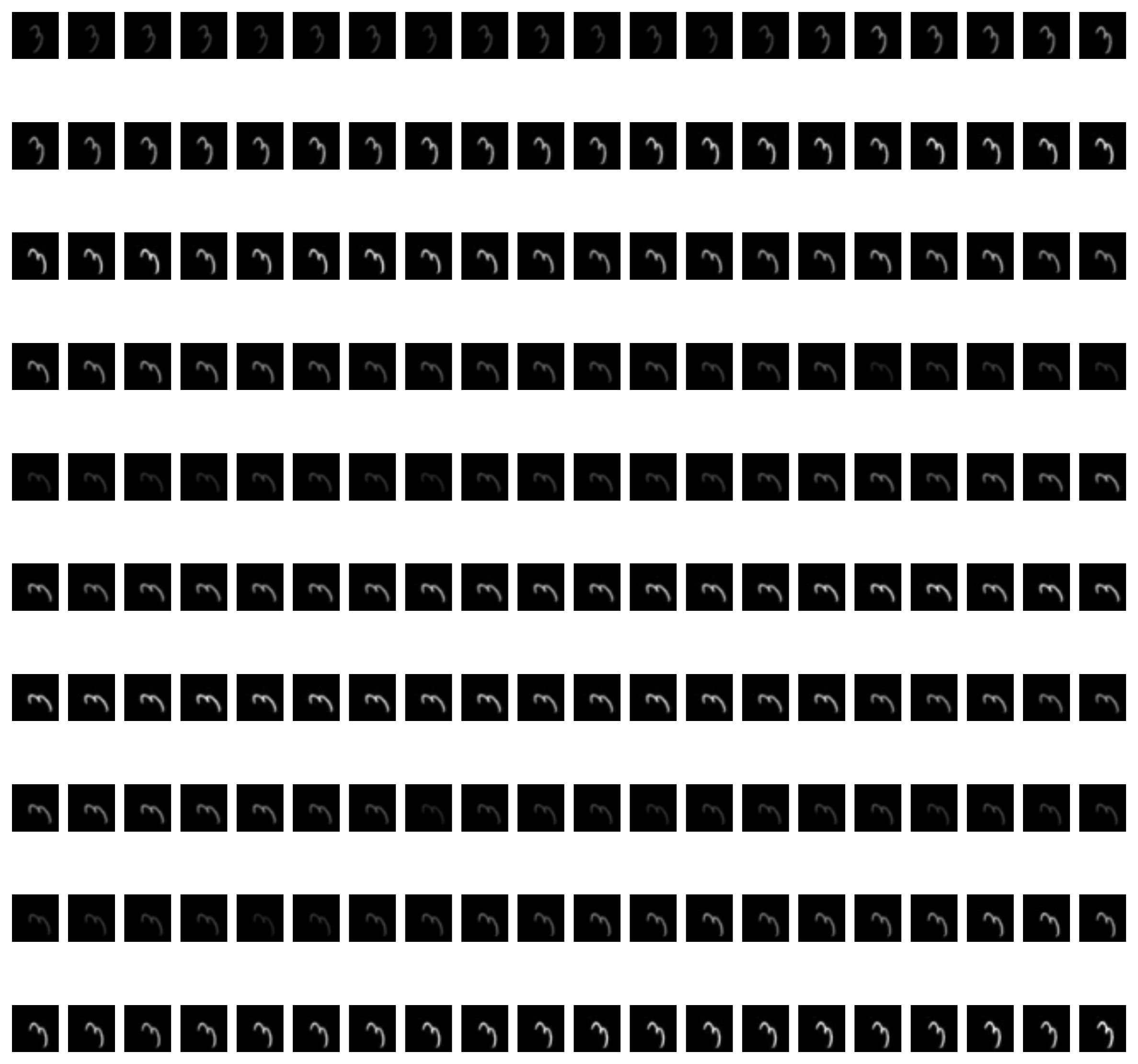}
  \caption{Visualisation of the modified digits $Y$.}
  \label{fig:generated_data}
\end{figure}

We created this dataset by taking a digit from the MNIST dataset and performing several manipulations to it. Each manipulated digit would become a measurement/observation instance ($Y$) and the corresponding values of the manipulations would become the auxiliary covariates ($X$). There were three main manipulations that were performed: a rotation about the centre of the digit, a translation (or shift) along the diagonal, and intensity of the digit (or contrast). Fig. \ref{fig:vis_aux} visualises the covariates $X$ for the temporal rotated digit dataset. Moreover, fig. \ref{fig:generated_data} visualises some sample data $Y$ that has been manipulated using the auxiliary covariates $X$.
\section{Pre-processing of clinical trial data}
\label{sec:suppl-clinical-data-description}

We make use of the Prostat\_Sanofi\_2007\_79 study (\url{https://data.projectdatasphere.org/projectdatasphere/html/content/79}). The study included a comparator arm with $371$ patients and several different measurement domains. We pre-processed the data to obtain data measurements $Y$ and auxiliary covariate information $X$. The measurement domains that we selected were laboratory measurements, demographic information, vital signs, adverse events, and concomitant medications. Moreover, we only chose observations where the start and end date of the adverse events and concomitant medications were known. We then transformed this data into a longitudinal format grouped by the unique patient IDs. The longitudinal samples $Y$ comprised of vital signs and laboratory measurements and the auxiliary covariates $X$ comprised of demographic information, adverse events, and concomitant medications. 
\begin{figure}[h]
\centering
  \includegraphics[width=0.5\linewidth]{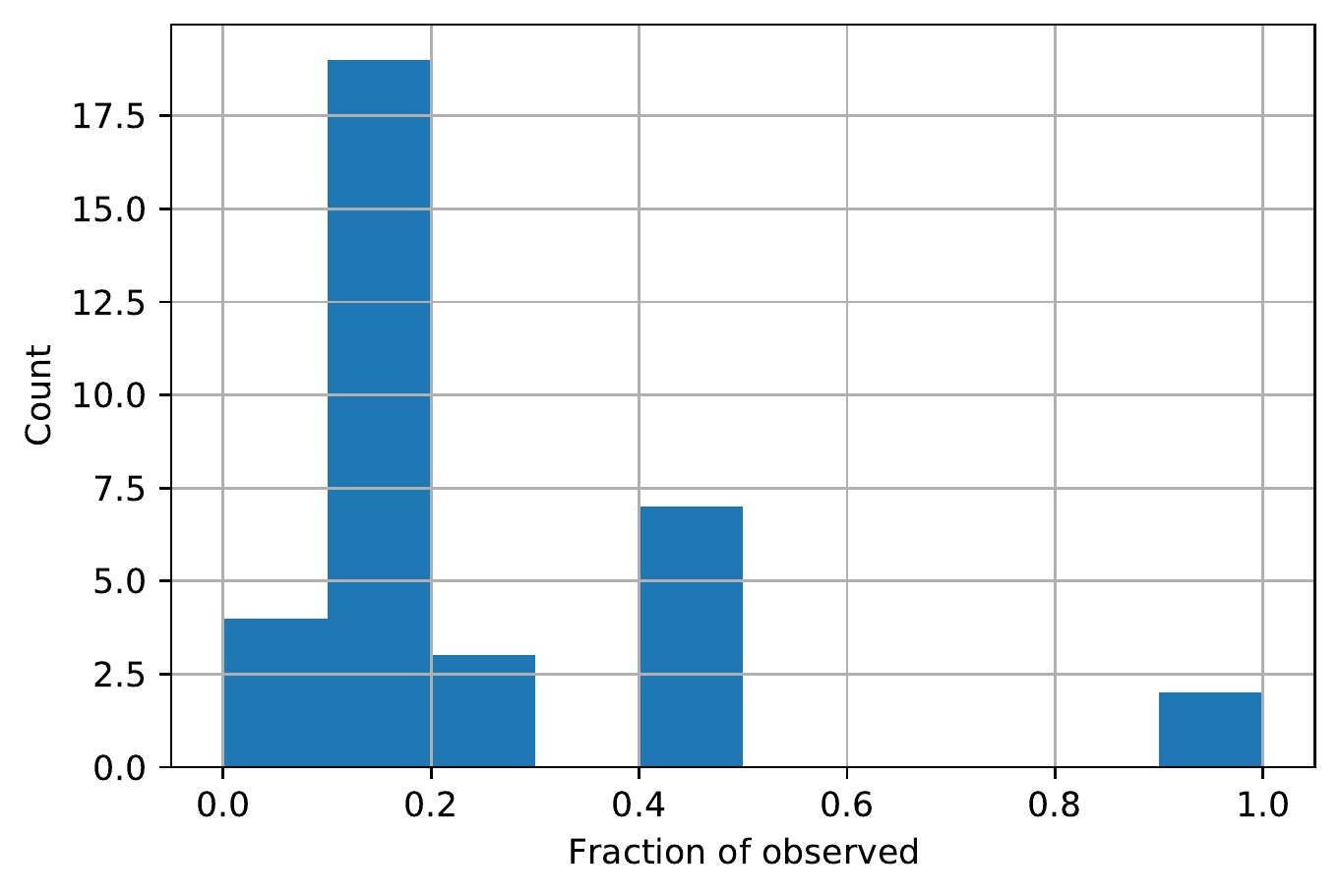}
  \caption{Histogram comparing the fraction of observed measurements for each measurement attribute in $Y$.}
  \label{fig:hist_observed_study1}
\end{figure}

We then computed the number of observed measurements in the dataset. Fig. \ref{fig:hist_observed_study1} visualises the fraction of observed measurements for each attribute of $Y$. About $25.01\%$ of the measurements in $Y$ were observed. In $Y$, we drop the columns with less than $10\%$ of measurements observed. Also, we keep the samples with at least $50\%$ of measurements observed and remove the rest. Moreover, we remove subjects that have less than $5$ samples. Finally, we perform min-max normalisation of the measurements $Y$.  We picked the 10 most occurring adverse events and 10 most occurring concomitant medications to include in $X$.
The training set comprised of $144$ individuals or $979$ observations, the validation set comprised of $20$ individuals or $141$ observations, and the test set comprised of $20$ individuals or $167$ observations. We experimented with different amounts of missing values (5\%, 10\%, 20\%, 30\%, and 40\%) artificially introduced into the auxiliary covariates of the training set, validation set, and test set.

\FloatBarrier
\section{Supplementary Tables}
\label{sec:supplementary_tables}
\begin{table*}[h]
\caption{Comparison of related VAE models.} \label{table:comparison}
\centering
\resizebox{0.8\textwidth}{!}{%
\begin{tabular}{lccccr}
\hline \\
\textbf{Models} & \begin{tabular}[c]{@{}c@{}}\textbf{Impute missing values}\\ \textbf{in measurements}\end{tabular} & \begin{tabular}[c]{@{}c@{}}\textbf{Impute missing values}\\ \textbf{in auxiliary covariates}\end{tabular} & \begin{tabular}[c]{@{}c@{}}\textbf{Used for}\\ \textbf{longitudinal modelling}\end{tabular} & \begin{tabular}[c]{@{}c@{}}\textbf{Conditional generative
}\\ \textbf{model}\end{tabular} & \textbf{References}\\
\hline
VAE    &        \cmark     &       \xmark     &    \xmark        &     \xmark     &     \citet{kingma2013auto}\\
CVAE   &        \cmark     &       \xmark     &    \xmark        &     \cmark     &      \citet{sohn2015learning}\\
HIVAE &        \cmark     &       \xmark      &    \xmark        &      \xmark    &      \citet{nazabal2020handling}\\
GP-VAE  &        \cmark     &       \xmark      &    \xmark        &      \cmark    &     \citet{fortuin2019multivariate}\\
L-VAE  &        \cmark     &       \xmark     &    \cmark        &      \cmark    &     \citet{ramchandran2021longitudinal}\\
\hline
CVAE with our method  &        \cmark     &       \cmark     &    \xmark        &      \cmark     &     This work\\
L-VAE with our method &        \cmark     &       \cmark     &    \cmark        &      \cmark    &      This work\\
\hline
\end{tabular}}
\end{table*}

\begin{table*}[h]
\caption{Comparison of related learning methods.} \label{table:comparison_learning}
\centering
\resizebox{0.6\textwidth}{!}{%
\begin{tabular}{lccc}
\hline \\
\textbf{Models} & \begin{tabular}[c]{@{}c@{}}\textbf{Impute missing values}\\ \textbf{in measurements}\end{tabular} & \begin{tabular}[c]{@{}c@{}}\textbf{Impute missing values}\\ \textbf{in auxiliary covariates}\end{tabular} & \begin{tabular}[c]{@{}c@{}}\textbf{Longitudinal and}\\ \textbf{temporal modelling}\end{tabular}\\
\hline
Mean imputation    &        \cmark     &       \xmark     &    \xmark   \\
KNN imputation   &        \cmark     &       \xmark     &    \xmark      \\
MIWAE \citep{mattei2019miwae} &        \cmark     &       \xmark      &    \xmark   \\
Masking   &    \xmark    & \xmark       &       \xmark \\
\hline
Our method  &        \cmark     &       \cmark     &    \cmark  \\
\hline
\end{tabular}}
\end{table*}

\begin{table*}[!h]
\centering
\caption{NLL values ($\times 10^2$) for predicting $Y$ in the test set  given a partially observed $X$ using CVAE based methods on Dataset 3 of the rotated digits dataset.}\label{tab:NLL_CVAE_dataset3}
\resizebox{0.8\textwidth}{!}{%
\begin{tabular}{lc|ccccr}
\hline
\multicolumn{1}{c}{\multirow{2}{*}{\bfseries Method}} & \multicolumn{1}{c|}{\multirow{2}{*}{\bfseries Dataset}} & \multicolumn{5}{c}{\bfseries Missing \%}                                                                  \\
\multicolumn{1}{c}{}                        & \multicolumn{1}{c|}{}                         & \bfseries 5\%               & \bfseries 10\%              & \bfseries 20\%              & \bfseries30\%             & \bfseries 40\%             \\ \hline
Vanilla CVAE                                                         & Dataset 3                                         & $0.58\pm 0.04$                     & $0.59\pm 0.05$                     & $0.58\pm 0.06$                     & $0.61\pm 0.05$                     & $0.69\pm 0.04$                     \\
CVAE with mean imputation                                            & Dataset 3                                                              & $0.57\pm 0.03$ & $0.58\pm 0.03$ & $0.57\pm 0.02$ & $0.6\pm 0.05$  & $0.67\pm 0.05$ \\
CVAE with KNN imputation                                             & Dataset 3                                         & $0.55\pm 0.04$                     & $0.58\pm 0.04$                     & $0.57\pm 0.03$                     & $0.59\pm 0.04$                     & $0.66\pm 0.03$                     \\
\rowcolor{Gray}
\textbf{CVAE with our method}                                                & Dataset 3                                                              & $\boldsymbol{0.52\pm 0.03}$                     & $\boldsymbol{0.54\pm 0.04}$                     & $\boldsymbol{0.55\pm 0.02}$                     & $\boldsymbol{0.57\pm 0.05}$                     & $\boldsymbol{0.6\pm 0.05}$                      \\
CVAE with oracle                                                               & Dataset 3                                         & $0.21\pm 0.01$                     & $0.32\pm 0.02$                     & $0.37\pm 0.02$                     & $0.41\pm 0.03$                     & $0.43\pm 0.02$                     \\ \hline
\end{tabular}}
\end{table*}

\begin{table}[!h]
\centering
\caption{MSE values for rotated digits dataset with both continuous and discrete-valued covariates.}\label{tab:MSE_het_cVAE}
\resizebox{\textwidth}{!}{%
\begin{tabular}{lc|ccccr}
\hline
\multicolumn{1}{c}{\multirow{2}{*}{\bfseries Method}} & \multicolumn{1}{c|}{\multirow{2}{*}{\bfseries Dataset}} & \multicolumn{5}{c}{\bfseries Missing \%}                                                                                \\
\multicolumn{1}{c}{}                        & \multicolumn{1}{c|}{}                         & \bfseries 5\%                               & \bfseries 10\%             & \bfseries 20\%             & \bfseries 30\%             & \bfseries 40\%             \\ \hline
Temporal L-VAE with mean impute                      & Dataset 3 (with discrete)                & $0.031\pm 0.005$                  & $0.037\pm 0.004$ & $0.044\pm 0.006$ & $0.051\pm 0.006$ & $0.062\pm 0.004$ \\
Temporal L-VAE with KNN impute                       & Dataset 3 (with discrete)                & $0.03\pm 0.004$                   & $0.035\pm 0.005$ & $0.041\pm 0.004$ & $0.045\pm 0.003$ & $0.059\pm 0.005$ \\
\rowcolor{Gray}
\textbf{Temporal L-VAE with our method}                       & Dataset 3 (with discrete)                & $\boldsymbol{0.018\pm 0.002}$                  & $\boldsymbol{0.021\pm 0.003}$ & $\boldsymbol{0.024\pm 0.004}$ & $\boldsymbol{0.035\pm 0.004}$ & $\boldsymbol{0.041\pm 0.003}$ \\
Temporal L-VAE with oracle                           & Dataset 3 (with discrete)                & $0.011\pm 0.001$ & $0.015\pm 0.002$ & $0.019\pm 0.001$ & $0.026\pm 0.003$ & $0.031\pm 0.004$ \\
\hline
\end{tabular}}
\end{table}

\begin{table}[!h]
\centering
\caption{NLL values $\times 10^3$ for rotated digits dataset with both continuous and discrete-valued covariates.}\label{tab:NLL_het_cVAE}
\resizebox{\textwidth}{!}{%
\begin{tabular}{lc|ccccr}
\hline
\multicolumn{1}{c}{\multirow{2}{*}{\bfseries Method}} & \multirow{2}{*}{\bfseries Dataset} & \multicolumn{5}{c}{\bfseries Missing \%}                                                                                                      \\
\multicolumn{1}{c}{}                                                 &                                                   & \bfseries 5\% & \bfseries 10\% & \bfseries 20\% & \bfseries 30\% & \bfseries 40\% \\ \hline
Temporal L-VAE with mean impute                                               & Dataset 3 (with discrete)                    & $1.9\pm 0.03$                & $2.6\pm 0.02$                 & $3.9\pm 0.06$                 & $4.8\pm 0.05$                 & $5.8\pm 0.06$                 \\
Temporal L-VAE with KNN impute                                                & Dataset 3 (with discrete)                    & $1.6\pm 0.02$                & $2.3\pm 0.03$                 & $3.6\pm 0.04$                 & $4.6\pm 0.05$                 & $5.4\pm 0.05$                 \\
\rowcolor{Gray}
\textbf{Temporal L-VAE with our method}                                                & Dataset 3 (with discrete)                    & $\boldsymbol{1.2\pm 0.02}$                & $\boldsymbol{1.8\pm 0.02}$                 & $\boldsymbol{2.4\pm 0.04}$                 & $\boldsymbol{3.7\pm 0.04}$                 & $\boldsymbol{4.8\pm 0.05}$                 \\ 
Temporal L-VAE with oracle                                                    & Dataset 3 (with discrete)                    & $0.95\pm 0.01$               & $1.3\pm 0.01$                 & $1.8\pm 0.03$                 & $3.1\pm 0.03$                 & $4.3\pm 0.04$                 \\
\hline
\end{tabular}}
\end{table}

\begin{table*}[!h]
\centering
\caption{Comparison of the imputation accuracy for covariates of the rotated digits dataset with both continuous and discrete-valued covariates.}\label{tab:MSE_label_disc_cov}
\resizebox{\textwidth}{!}{%
\begin{tabular}{l|c|c|c|c|c}
\hline
\multicolumn{1}{c|}{\multirow{3}{*}{\bfseries Method}} & \multicolumn{5}{c}{\bfseries Missing \%} \\ 
\multicolumn{1}{c|}{} & \bfseries 5\% & \bfseries 10\% & \bfseries 20\% & \bfseries30\% & \bfseries 40\%  \\
\multicolumn{1}{c|}{} & \begin{tabular}{lr} \bfseries MSE $\downarrow$ & \bfseries Acc.\% $\uparrow$ \end{tabular} & \begin{tabular}{lr}\bfseries MSE $\downarrow$ & \bfseries Acc.\% $\uparrow$\end{tabular} & \begin{tabular}{lr}\bfseries MSE $\downarrow$ & \bfseries Acc.\% $\uparrow$\end{tabular} & \begin{tabular}{lr}\bfseries MSE $\downarrow$ & \bfseries Acc.\% $\uparrow$\end{tabular} & \begin{tabular}{lr}\bfseries MSE $\downarrow$ & \bfseries Acc.\% $\uparrow$\end{tabular} \\ \hline

Mean impute   & \begin{tabular}{lr}  $0.65 $ & $72\% $ \end{tabular}  &   \begin{tabular}{lr}  $0.68$ & $70\% $ \end{tabular} & \begin{tabular}{lr}  $0.71$ & $68\%$ \end{tabular}           &        \begin{tabular}{lr}  $0.73$ & $67\% $\end{tabular}    &         \begin{tabular}{lr} $ 0.76 $ & $64\%$ \end{tabular}                         \\

KNN impute  &  \begin{tabular}{lr}  $0.62$& $74\%$ \end{tabular}   & \begin{tabular}{lr}  $0.66$ & $72\% $ \end{tabular}  &  \begin{tabular}{lr}  $0.68$& $71\%$ \end{tabular}    &  \begin{tabular}{lr} $ 0.7$ & $69\%$\end{tabular}  &  \begin{tabular}{lr}  $0.72$& $66\%$ \end{tabular}                              \\
\rowcolor{Gray}
\textbf{L-VAE with our method}  &   \begin{tabular}{lr} $\boldsymbol{0.58 \pm 0.1} $ & $ \boldsymbol{79\% \pm 1.2 }$\end{tabular}  &     \begin{tabular}{lr} $ \boldsymbol{0.61 \pm 0.1} $ & $ \boldsymbol{76\% \pm 1.5}$ \end{tabular}  & \begin{tabular}{lr} $ \boldsymbol{0.62 \pm 0.2}$ & $ \boldsymbol{75\% \pm 2.6 }$\end{tabular}    &      \begin{tabular}{lr}  $\boldsymbol{0.65 \pm 0.3}$ & $\boldsymbol{73\% \pm 2.2} $\end{tabular} &  \begin{tabular}{lr}  $\boldsymbol{0.67 \pm 0.3}$ & $\boldsymbol{71\% \pm 3.8}$  \end{tabular} \\ \hline                                  
\end{tabular}}
\end{table*}

\begin{table*}[!h]
\centering
\caption{Comparison of the imputation accuracy for covariates of the clinical trial dataset}\label{tab:MSE_label_clinical}
\resizebox{\textwidth}{!}{%
\begin{tabular}{l|c|c|c|c|c}
\hline
\multicolumn{1}{c|}{\multirow{3}{*}{\bfseries Method}} & \multicolumn{5}{c}{\bfseries Missing \%} \\ 
\multicolumn{1}{c|}{} & \bfseries 5\% & \bfseries 10\% & \bfseries 20\% & \bfseries30\% & \bfseries 40\%  \\
\multicolumn{1}{c|}{} & \begin{tabular}{lr} \bfseries MSE $\downarrow$ & \bfseries Acc.\% $\uparrow$ \end{tabular} & \begin{tabular}{lr}\bfseries MSE $\downarrow$ & \bfseries Acc.\% $\uparrow$\end{tabular} & \begin{tabular}{lr}\bfseries MSE $\downarrow$ & \bfseries Acc.\% $\uparrow$\end{tabular} & \begin{tabular}{lr}\bfseries MSE $\downarrow$ & \bfseries Acc.\% $\uparrow$\end{tabular} & \begin{tabular}{lr}\bfseries MSE $\downarrow$ & \bfseries Acc.\% $\uparrow$\end{tabular} \\ \hline

Mean impute   & \begin{tabular}{lr}  $4.7 $ & $61\% $ \end{tabular}  &   \begin{tabular}{lr}  $5.3$ & $59\% $ \end{tabular} & \begin{tabular}{lr}  $5.7$ & $57\%$ \end{tabular}           &        \begin{tabular}{lr}  $6.1$ & $56\% $\end{tabular}    &         \begin{tabular}{lr} $ 6.4 $ & $55\%$ \end{tabular}                         \\

KNN impute  &  \begin{tabular}{lr}  $4.4$& $70\%$ \end{tabular}   & \begin{tabular}{lr}  $5.1$& $68\% $ \end{tabular}  &  \begin{tabular}{lr}  $5.3$& $67\%$ \end{tabular}    &  \begin{tabular}{lr} $ 5.5$ & $65\%$\end{tabular}  &  \begin{tabular}{lr}  $5.9$& $62\%$ \end{tabular}                              \\
\rowcolor{Gray}
\textbf{L-VAE with our method}  &   \begin{tabular}{lr} $\boldsymbol{3.9 \pm 0.2} $ & $ \boldsymbol{78\% \pm 1.8 }$\end{tabular}  &     \begin{tabular}{lr} $ \boldsymbol{4.4 \pm 0.2} $ & $ \boldsymbol{73\% \pm 2.5}$ \end{tabular}  & \begin{tabular}{lr} $ \boldsymbol{4.6 \pm 0.3}$ & $ \boldsymbol{72\% \pm 3.1 }$\end{tabular}    &      \begin{tabular}{lr}  $\boldsymbol{4.8 \pm 0.4}$ & $\boldsymbol{69\% \pm 3.6} $\end{tabular} &  \begin{tabular}{lr}  $\boldsymbol{5.3 \pm 0.4}$ & $\boldsymbol{67\% \pm 4.1}$  \end{tabular} \\ \hline                              
\end{tabular}}
\end{table*}

\begin{table}[!h]
\begin{center}
\begin{adjustbox}{width=0.8\textwidth}
\begin{tabular}{ c l r } 
\hline
 & Hyperparameter & Value \\
\hline
\multirow{12}{6em}{Inference network} & Dimensionality of input & $36 \times 36 + 4$ \\ 
& Number of convolution layers & 2 \\ 
& Number of filters per convolution layer & 144 \\ 
& Kernel size &  $3 \times 3$\\
& Stride &  2\\
& Pooling & Max pooling\\
& Pooling kernel size & $2 \times 2$\\
& Pooling stride & 2\\
& Number of feedforward layers & 2 \\
& Width of feedforward layers & 300, 30 \\
& Dimensionality of latent space & $L$\\
& Activation function of layers & RELU \\
\hline
\multirow{8}{6em}{Generative network} & Dimensionality of input & $L$ \\
& Number of transposed convolution layers & 2 \\
& Number of filters per transposed convolution layer & 256 \\
& Kernel size &  $4 \times 4$\\
& Stride &  2\\
& Number of feedforward layers & 2 \\
& Width of feedforward layers & 30, 300 \\
& Activation function of layers & RELU \\
\hline
\end{tabular}
\end{adjustbox}
\end{center}
\caption{Neural network architectures used in the rotated digits dataset.}
\label{table:nnet_spec_healthMNIST}
\end{table}

\begin{table}[!h]
\begin{center}
\begin{adjustbox}{width=0.8\textwidth}
\begin{tabular}{ c l r } 
\hline
 & Hyperparameter & Value \\
\hline
\multirow{5}{6em}{Inference network} & Dimensionality of input & 28 + 13  \\ 
& Number of feedforward layers & 2 \\
& Number of elements in each feedforward layer & 128, 64 \\
& Dimensionality of latent space & $L$\\
& Activation function of layers & RELU \\
\hline
\multirow{4}{6em}{Generative network} & Dimensionality of input & $L$ \\
& Number of feedforward layers & 2 \\
& Number of elements in each feedforward layer & 64, 128 \\
& Activation function of layers & RELU \\
\hline
\end{tabular}
\end{adjustbox}
\end{center}
\caption{Neural network architectures used in the clinical trial dataset.}
\label{table:nnet_spec_physionet}
\end{table}


\end{document}